\documentclass{article}

\PassOptionsToPackage{numbers,sort&compress}{natbib}
\usepackage[preprint]{neurips_2026}

\usepackage[utf8]{inputenc}
\usepackage[T1]{fontenc}
\usepackage{hyperref}
\usepackage{url}
\usepackage{booktabs}
\usepackage{longtable}
\usepackage{multirow}
\usepackage{amsfonts}
\usepackage{amsmath}
\usepackage{nicefrac}
\usepackage{microtype}
\usepackage{xcolor}
\usepackage{colortbl}
\definecolor{tabheader}{HTML}{D9E5F2}
\definecolor{tabrowalt}{HTML}{F2F4F8}
\usepackage{graphicx}
\usepackage{enumitem}
\usepackage{caption}
\usepackage{subcaption}
\usepackage{wrapfig}
\usepackage{mathptmx}
\hypersetup{hidelinks}

\title{GHGbench: A Unified Multi-Entity, Multi-Task Benchmark for Carbon Emission Prediction}

\author{%
  Yifan Duan \quad Siyuan Zheng \quad Lihuan Li \quad Chao Xue \quad Flora Salim \\
  School of Computer Science and Engineering \\
  University of New South Wales \\
  \texttt{yifan.duan3@unsw.edu.au} \quad \texttt{flora.salim@unsw.edu.au}
}

\begin{document}

\maketitle

\begin{abstract}
Open datasets and benchmarks for entity-level carbon-emission prediction
remain fragmented across access, scale, granularity, and evaluation.
We introduce \textbf{GHGbench}, an open dataset and benchmark for
company- and building-level greenhouse-gas prediction. The company
track contains 32{,}000+ company-year records from 12{,}000+ firms with
Scope~1+2 and Scope~3 disclosures and financial/sectoral signals; the
building track harmonises 491{,}591 building-year records from 13 open
sources into a single schema across 26 metropolitan areas (10 U.S.,
15 Australian, 1 Singaporean), with climate covariates and multimodal
remote-sensing embeddings. GHGbench defines canonical splits with in-distribution
and cross-region/city transfer as \emph{primary} tasks and temporal
hold-out plus short-horizon forecasting as \emph{supplementary}
appendix evidence; headline baselines span gradient-boosted trees, a
tabular foundation model, MLP, FT-Transformer, and multimodal fusion,
with an LLM panel as \emph{auxiliary}, all evaluated under multi-seed
paired-bootstrap tests. Three benchmark-level
findings emerge: (i) \emph{building emissions are structurally harder
than company emissions}; (ii) \emph{the in-distribution to
out-of-distribution gap dwarfs any within-model gap} across both the
company track and the building track, and a tabular foundation model
is, to our knowledge, the first baseline to open a
paired-bootstrap-significant gap over tuned trees on a multi-city
building-emissions task; (iii) \emph{multimodal remote-sensing
embeddings help precisely where tabular generalisation breaks}. GHGbench also exposes
catastrophic city transfer and the sector-factor lookup ceiling as
systematic failure modes. Code and
reconstruction recipes are available at
\href{https://github.com/cruiseresearchgroup/GHGbenchmark}{\textbf{\textcolor{blue}{GHGbench}}}.
\end{abstract}

\section{Introduction}
\label{sec:intro}

Global commitments to net-zero greenhouse-gas (GHG) emissions have made quantitative emission prediction a practical requirement for climate policy~\citep{ipcc2023ar6}, finance~\citep{bolton2021carbon}, and urban operations~\citep{creutzig2015urban}. Although emissions accounting has long been a policy concern~\citep{wri2004ghgprotocol}, machine-learning-based entity-level prediction is still an emerging area~\citep{rolnick2022tackling}, and the open benchmarking ecosystem has not caught up~\citep{nguyen2023scope3}. Company-level studies often rely on licensed disclosure or financial feeds~\citep{nguyen2021predicting,serafeim2022scope3}, while building-level data remain scattered across city portals with incompatible schemas~\citep{yap2025multicity,li2024genbenchmark}. As a result, it remains difficult to compare models on open, reproducible, entity-level tasks that reflect the geographic, temporal, and schema shifts encountered in real emissions-accounting workflows.

Existing approaches cluster into three lines of work, each with structural limits. Company-level studies show that financial, sectoral, and disclosure features can predict Scope~1/2 and Scope~3 emissions, but they often depend on licensed feeds such as CDP, Bloomberg, Refinitiv, or Compustat~\citep{nguyen2021predicting,han2021estimation,serafeim2022scope3,nguyen2023scope3}. Building-energy and operating-carbon benchmarks provide valuable data, but are typically limited by geography, incompatible public schemas, or non-reusable splits~\citep{miller2020bdg2,emami2023buildingsbench,yap2025multicity,li2024genbenchmark}. Sector-level emission-factor resources support enrichment, but aggregate above the firms and buildings where predictions are needed~\citep{stadler2018exiobase,guo2024exioml,guo2025green}. These limitations motivate a benchmark that evaluates company and building emissions under shared tasks, canonical splits, metrics, and uncertainty reporting.

\begin{figure}[t]
  \centering
  \includegraphics[width=\linewidth]{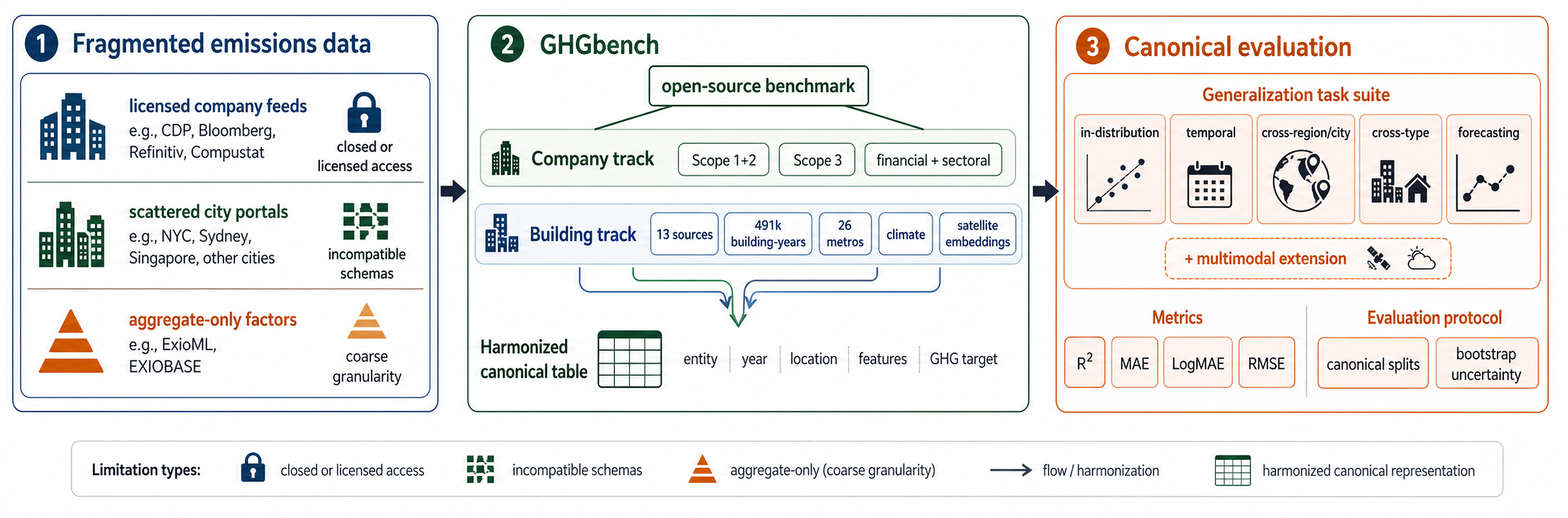}
  \caption{GHGbench overview. Left: fragmented emissions resources. Middle: harmonised company + building records with climate and satellite signals. Right: canonical tasks, metrics, and uncertainty protocols.}
  \label{fig:intro_overview}
  \vspace{-0.5em}
\end{figure}

To address this gap, we introduce \textbf{GHGbench}, \emph{to our knowledge the
first open dataset and benchmark to jointly evaluate company- and building-level
carbon-emission prediction under a shared task suite that spans in-distribution
regression, cross-region/city transfer, short-horizon forecasting, and
multimodal fusion, with paired-bootstrap uncertainty reporting
on the headline structured-baseline comparisons.}
The company track evaluates Scope~1+2 and Scope~3 emissions from corporate
disclosures enriched with financial and sectoral signals, released through
reconstruction recipes that respect source-data terms. The building track
harmonises 491{,}591 building--year records from 13 open sources across 26
metropolitan areas in the United States, Australia, and Singapore, with unified
units, property types, coordinates, and annual GHG targets. Across both tracks,
GHGbench provides canonical splits, shared metrics, and uncertainty reporting,
while auxiliary weather and satellite-derived
signals~\citep{nasa_power_daily_api,drusch2012sentinel2} support multimodal
evaluation. Figure~\ref{fig:intro_overview} summarises this flow.

Specifically, the company track starts from public corporate greenhouse-gas
disclosures: we normalise identifiers and units, construct Scope~1+2 and
Scope~3 panels with deduplication and outlier filtering, and enrich each
company-year with financial fields, sectoral emission factors,
sub-sector target encoding, and business-summary text for LLM baselines,
producing matched wide and strict feature regimes.
The building track harmonises 13 public disclosure sources into one
schema with a canonical 22-category property taxonomy, per-building-year
NASA POWER~\citep{nasa_power_daily_api} climate covariates, geocoded
coordinates, and Sentinel-2~\citep{drusch2012sentinel2} + Clay multimodal
embeddings, yielding 491{,}591 building-year rows from 100{,}984
buildings. Both tracks share a multi-task suite
spanning in-distribution regression, cross-region/city transfer, temporal
hold-out, cross-property-type stress testing, and short-horizon
forecasting; we evaluate tuned gradient-boosted trees,
TabPFN~v2~\citep{hollmann2025tabpfn}, MLP, FT-Transformer~\citep{gorishniy2021ftt}, time-series
foundation models~\citep{ansari2024chronos,das2024timesfm,woo2024moirai},
and an LLM panel under multi-seed runs, bootstrap confidence intervals,
and 1000-resample paired bootstrap on shared test rows.

Three benchmark-level findings emerge. (i) \emph{Building emissions
are structurally harder than company emissions}: companies are
dominated by firm size, while building emissions also depend on
occupancy, runtime, and behaviour that public disclosure does not
capture. (ii) \emph{The in-distribution to out-of-distribution gap
dwarfs any within-model gap on both tracks}; on the harder building
OOD setting, a pretrained tabular foundation model is, to our
knowledge, the first baseline to open a paired-bootstrap-significant
gap over tuned trees on a real-world, multi-city, multi-source
benchmark. (iii) \emph{Multimodal remote-sensing embeddings help
precisely where tabular generalisation breaks}: a no-op
in-distribution, measurable gain on cross-city transfer. The benchmark also exposes
systematic failure modes (catastrophic city transfer and the
sector-factor lookup ceiling) that mark where deep, multimodal, and
foundation models must improve next.

Our contributions are: (i) an open dataset and benchmark with 32{,}000+ company-year records reconstructable from free public APIs and 491{,}591 harmonised building-year records across 26 metros in three countries; (ii) a multi-task suite with canonical splits for in-distribution, temporal, cross-region/city, cross-property-type, and forecasting evaluation, aligned climate and remote-sensing signals, and a paired-bootstrap protocol; (iii) three benchmark-level findings (cross-entity difficulty asymmetry, ID-to-OOD gap dominating within-model variation, and multimodal embeddings helping where tabular generalisation breaks); and (iv) an open-source release with code, data, and a one-command quickstart kit.

\section{Related Work}
\label{sec:related}

\noindent\textbf{Company-level emissions prediction and ESG disclosure.}
Machine-learning approaches to corporate emissions estimation have largely focused on company-level financial, sectoral, and ESG features. Prior work estimates Scope~1+2 emissions from firm attributes and disclosure-linked financial data~\citep{nguyen2021predicting,han2021estimation,assael2023greenhouse}, extends prediction to Scope~3 categories~\citep{ghgprotocol_scope3,serafeim2022scope3}, and documents the large divergence among commercial Scope~3 providers~\citep{nguyen2023scope3}. A related line estimates emissions through sector classification: EXIOBASE input--output factors~\citep{stadler2018exiobase} are packaged as ML-ready benchmarks in ExioML~\citep{guo2024exioml}, while GREEN/ExioNAICS~\citep{guo2025green} maps free-text company descriptions to NAICS sectors. Corporate climate text has also become an NLP target, including ClimateBERT~\citep{webersinke2021climatebert}, GHG-extraction benchmarks~\citep{beck2025ghgextraction}, and open LLM-finetuning corpora~\citep{nopperl2024corporate}. These studies typically isolate one target family, one data source, or one text-extraction task; GHGbench's company track instead evaluates Scope~1+2, Scope~3, text/LLM baselines, and short-horizon forecasting under one protocol, alongside a building-level benchmark.

\noindent\textbf{Building energy and operating-carbon benchmarks.}
Building-level energy prediction and benchmarking provide the closest analogues for GHGbench's building track~\citep{deng2018predictive,arjunan2020energystarpp}. The Building Data Genome Project~2~\citep{miller2020bdg2} releases hourly meter data from 1{,}636 non-residential buildings, and BuildingsBench~\citep{emami2023buildingsbench} adds a large simulated stock with zero-shot/transfer-learning tasks for short-term load forecasting. Closer to annual operating carbon, \citet{li2024genbenchmark} benchmark twelve U.S.\ cities under a single national schema, while \citet{yap2025multicity} combine disclosures from five cities (New~York, Seattle, Washington~DC, Melbourne, Singapore) with multimodal geospatial inputs and graph deep learning. Satellite and rooftop imagery have been used as auxiliary signals for building-level energy prediction~\citep{streltsov2020overhead,dougherty2021schmear}, and geospatial encoders such as SatMAE, Scale-MAE, Prithvi-EO, and Clay~\citep{cong2022satmae,reed2023scalemae,szwarcman2024prithvi} make Sentinel-2~\citep{drusch2012sentinel2} embeddings readily usable. Unlike BuildingsBench~\citep{emami2023buildingsbench}, which targets short-term load forecasting on simulated meter data, GHGbench targets annual operating-carbon prediction on real disclosures, and differs from prior building-emissions resources along four axes that together define the benchmark gap. \textbf{(i) Coverage}: 26 metros across the U.S., Australia, and Singapore, adding 15 Australian cities under a NABERS/BEEC schema harmonised with U.S.\ ENERGY STAR Portfolio Manager disclosures. \textbf{(ii) Evaluation protocol}: canonical building-grouped, leave-one-city-out, leave-one-property-type-out, temporal hold-out, and one-step forecasting splits released as the benchmark contract, rather than a single in-distribution regression setting. \textbf{(iii) Statistical rigour}: every headline model-ordering claim is accompanied by a 1000-resample paired bootstrap on shared test rows. \textbf{(iv) Multimodal scope}: aligned NASA POWER climate covariates and Sentinel-2 + Clay embeddings released as a transfer-augmenting signal evaluated separately on in-distribution and cross-city splits.

\noindent\textbf{Cross-entity evaluation and modern baselines.}
Across both entity levels, the central open question is not only which model performs best in distribution, but how model behaviour changes under schema, regional, temporal, property-type, and modality shifts. Tree ensembles remain strong in typical in-distribution tabular regimes~\citep{grinsztajn2022tabular}, and TableShift~\citep{gardner2023tableshift} provides a general distribution-shift benchmark for tabular classification. Modern baselines now span TabPFN~v2 for tabular prediction~\citep{hollmann2025tabpfn}, Chronos, TimesFM, and Moirai for time-series forecasting~\citep{ansari2024chronos,das2024timesfm,woo2024moirai}, and language models for disclosure understanding~\citep{webersinke2021climatebert,nopperl2024corporate,guo2025green}. GHGbench uses these as evaluation targets, asking how classical, neural, language, time-series, and satellite-augmented models behave when the data are heterogeneous, multimodal, and explicitly split to test generalisation.

\section{Dataset}
\label{sec:dataset}

GHGbench is organised around two entity levels that are central to emissions
accounting but rarely benchmarked under a shared protocol: companies and
buildings. The company track uses annual corporate disclosures to evaluate
company-level Scope~1+2 and Scope~3 prediction. The building track harmonises
public building-energy and emissions disclosures into an annual building-level
operating-carbon benchmark. Both tracks are built from noisy real-world disclosures and augmented
with structured external signals, so models face the data constraints
of real emissions-accounting systems. The
building track is released as a fully open dataset; the company track ships
as a one-command reconstruction script that pulls disclosures via a free
public API.

\begin{figure}[t]
  \centering
  \includegraphics[width=\linewidth]{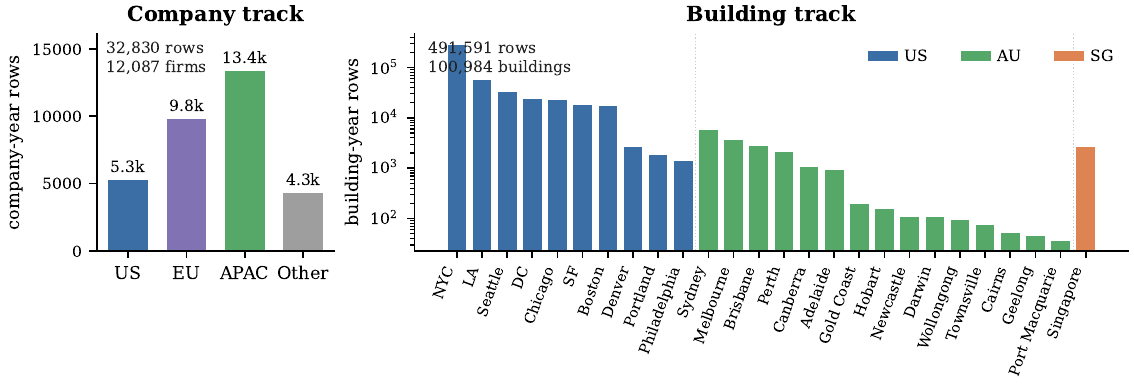}
  \caption{Dataset coverage. Left: company-year rows by region. Right: building-year rows by metropolitan area, grouped by country.}
  \label{fig:dataset_coverage}
\end{figure}

\subsection{Company Track}
\label{sec:dataset_t1}

The company track starts from disclosure records collected by the
Climate Data Utility (CDU),\footnote{\url{https://www.climatedatautility.org/}; free registration provides full API access.}
which provides annual company-level greenhouse-gas disclosures from
2018 to 2023. We normalise company identifiers and country codes, convert disclosure
fields to numeric tonnes CO$_2$e, construct Scope~1+2 from reported Scope~1 and
location-based Scope~2, deduplicate company--year records, and apply
paper-facing outlier filters. The resulting enriched panel contains 32{,}830
company--year rows from 12{,}087 companies, with 31{,}331 usable Scope~1+2
labels. Scope~3 is joined from the raw CDU file at run time, with
18{,}763 lenient-coverage rows and 9{,}814 strict-coverage rows after
row-level filtering (these are Scope~3 row regimes, distinct from the
Scope~1+2 feature regimes defined below).

Each company-track record combines country, year, and sector metadata with
financial and text signals. Public-company enrichment uses name-to-ticker
matching, cached Yahoo-Finance financial statements, company business summaries
as text inputs for LLM baselines, and a foreign-exchange correction for local-currency
financial fields. Sectoral emissions are supplied by an
ExioML/EXIOBASE-derived lookup over GICS sector, ExioML region, and reporting
year~\citep{stadler2018exiobase,guo2024exioml}, with SICS sub-sector target
encoding as an additional sectoral signal; an ExioNAICS/GREEN-style
sector-text linkage is provided as a complementary classification resource
only~\citep{guo2025green}. The company track exposes four feature
configurations used as the column axis in the headline experiments:
\emph{Open} (country, sector, year, and the ExioML factor on the full
panel), \emph{Open matched} (the same features restricted to the
yfinance-matched subset for an apples-to-apples comparison), \emph{+firm}
(Open matched plus revenue, employees, market capitalisation, and EBITDA),
and \emph{+SICS} (+firm with SICS sub-sector target encoding). Filtering
rules are detailed in Appendix~\ref{app:data_harmonisation}. Because the
company track spans 2018--2023, its forecasting task is necessarily
short-horizon.

\subsection{Building Track}
\label{sec:dataset_t2}

The building track contains 491{,}591 building--year records from 13 public
building-performance sources in the United States, Australia, and Singapore:
100{,}984 buildings, 26 metro-level cities (10 U.S., 15 Australian, 1 Singaporean), and 471{,}070 rows with annual
operating-GHG targets. The raw year span is 2011--2026, with broad multi-city
coverage strongest through 2024. All sources are mapped into one schema for
city, building identifier, year, property type, floor area, energy intensity,
rating score, GHG emissions, energy use, coordinates, and coordinate
provenance.

The harmonisation layer standardises units, collapses duplicate
building--years, removes or blanks physically invalid values, and maps raw
property strings into a 22-category taxonomy. It also resolves source-specific
issues that affect model comparability: Australian suburb/state labels are
mapped to 13 metro-level cities, NABERS/BEEC annual consumption is converted
from MJ to kBtu (1\,MJ = 0.9478\,kBtu) before computing EUI, and source-provided
coordinates are retained where available. For sources that publish addresses
but not coordinates, we run building-level geocoding against Nominatim
(OpenStreetMap), rate-limited to 1 request per second with a resumable cache
and de-duplicated by query. In the canonical table this supplies
79{,}155 geocoded building--year rows, corresponding to 20{,}511 unique
buildings across Los Angeles, Boston, Singapore, Denver, and Portland; each
match is filtered by a city-radius sanity check before merging with explicit
coordinate provenance.
Denver, Philadelphia, and Portland have sparse reporting histories and
are excluded from temporal and forecasting evaluations while retained
in all other experiments. Additional
source-specific rules appear in Appendix~\ref{app:data_harmonisation}.

\textbf{Feature tiers.} Because public disclosures expose different
fields across countries, building features fall into nine \emph{feature
tiers} organised in three ladders. A 26-city \emph{cross-country}
ladder uses only fields available in every source (size, coordinates,
year, degree days). A \emph{U.S.}\ ladder progressively adds property
type, year built, ENERGY~STAR score, source-EUI, and direct energy use
as cities permit; an \emph{Australian} ladder adds site EUI and
NABERS-derived rating tiers. Each tier is labelled \emph{clean},
\emph{proxy-rich}, or \emph{direct-energy-proxy} by whether its features
include target-correlated proxies, and headline numbers are reported
within tier since cross-tier differences reflect richer disclosures
rather than model superiority. The full registry is in
Appendix~\ref{app:feature_tiers}; for cross-type evaluation, the
22-category taxonomy is further collapsed into seven coarse property
groups (Appendix~\ref{app:data_harmonisation}).\label{sec:feature_tiers}

\subsection{External Alignment}
\label{sec:dataset_external}

The building track is augmented with building-year climate features from the
NASA POWER daily point API~\citep{nasa_power_daily_api}, aggregated into degree
days, annual mean temperature, relative humidity, solar radiation, and wind
speed for 491{,}416 building--year rows. The multimodal extension aligns building-track rows
to Sentinel-2 imagery~\citep{drusch2012sentinel2} and Clay image
embeddings.\footnote{Clay v1.5:
\url{https://github.com/Clay-foundation/model}.} Headline multimodal
experiments use 369{,}698 rows with valid building-footprint patches and
target labels; tabular and tabular-plus-Sentinel experiments share this
eligible subset. We compare three S2 feature variants: raw
1024-dimensional Clay embeddings, PCA-64, and PCA-128, all fitted on
training rows only.

\section{Benchmark Design}
\label{sec:benchmark}

GHGbench specifies a task suite, evaluation splits, metrics, and baseline families rather than a single leaderboard. This section formalises the evaluation rules used in the experiments; data and feature definitions (including the building feature tiers) are in Section~\ref{sec:dataset}.

\subsection{Task Suite}
\label{sec:benchmark_tasks}

GHGbench operationalises four evaluation capabilities rarely tested jointly. These capabilities operationalise the three benchmark-level questions from \S\ref{sec:intro}: cross-entity difficulty (in-distribution on both tracks), the ID-to-OOD gap (cross-distribution transfer), and modality lift under shift (multimodal $\times$ cross-city). Two primary capabilities (in-distribution prediction, cross-distribution transfer) frame the main results; two supplementary axes (temporal drift, one-step forecasting) appear in the appendix. The full task index is in Appendix~Table~\ref{tab:tasks}.

\textbf{In-distribution and unseen-entity accuracy.} How well do models fit training-distribution data, and how does accuracy hold up on entities unseen during training? Company Scope~1+2 and Scope~3 regression use stratified random splits; building emissions use two split variants---\emph{row-random} (rows shuffled at row granularity; sensitivity analysis only) and \emph{building-grouped} (rows partitioned by \texttt{building\_id} so every test building is held out from train; the main-paper deployable setting, since real deployments score on entities unseen during training).

\textbf{Cross-distribution transfer.} How much do predictions degrade when training and evaluation distributions diverge? At the company level, we leave one of \{U.S., EU, APAC\} out at a time, training on the other two and evaluating on the third. At the building level, we hold out one of 26 metros at a time; spanning the U.S., Australia, and Singapore, this also exposes country and schema shift. As an appendix-level stress test, we additionally hold out one of seven property categories at a time (the 22-category taxonomy collapsed as described in \S\ref{sec:dataset_t2}).

\textbf{Supplementary capabilities.} \emph{Temporal drift} asks whether models trained on past years generalise forward when reporting cohorts and energy mixes change; the building-level hold-out trains on years $\le 2019$, validates on 2020, and tests on years $\ge 2021$ (Appendix~\ref{app:task_b_grid}). \emph{Forecasting} asks whether lagged context alone suffices for future-period prediction; the company panel forecasts 2022 emissions from 2018--2021 context, while the building panel evaluates each test year $y\!\in\!\{2021,2022,2023,2024\}$ separately as horizon $h\!=\!y\!-\!2019$ from a fixed train cut at 2019, exposing horizon decay (Appendix~Table~\ref{tab:app_task_e_horizon}). Both axes use a short five-year panel and overlap with the temporal hold-out, so they are reported as appendix-level evidence (Appendix~\ref{app:t2_transfer_forecasting}).

\subsection{Metrics and Reporting Convention}
\label{sec:benchmark_metrics}

Company-track baselines are evaluated in $\log_{10}$ space, where we report log-space MAE, RMSE, $R^2$, and Pearson correlation, plus raw-scale percentage errors derived after exponentiating predictions. Building-track baselines are evaluated in raw tonnes CO$_2$e, where we report MAE, RMSE, MAPE, $R^2$, NRMSE, and LogMAE. Headline tables and figures use $R^2$ together with MAE or LogMAE, depending on whether the comparison emphasises raw-scale interpretability or heavy-tail robustness; we do not use MAPE as a standalone headline metric because near-zero emissions inflate it disproportionately even with the standard $\varepsilon$-stabilised denominator.

\subsection{Statistical Protocol}
\label{sec:benchmark_stats}

Every headline number is reported as the mean over three seeds with bootstrap 95\% confidence intervals (1000 resamples of the test rows). Pairwise model comparisons on a shared task and feature configuration use a \emph{paired bootstrap} on the same test rows: for each of 1000 resamples we recompute both models' losses on the resampled indices and record the per-resample loss difference, yielding a two-sided $p$-value on $\Delta$ and a 95\% CI. We call a gap \emph{paired-bootstrap-significant} when the resulting CI excludes zero (equivalently $p<0.05$). This is the basis for the ID-to-OOD gap and TabPFN-vs-trees claims in \S\ref{sec:intro} and \S\ref{sec:experiments}. Concrete split parameters (stratification, train/val/test ratios, year cuts, excluded cities) are listed in Appendix~\ref{app:splits}.

\subsection{Baselines}
\label{sec:benchmark_baselines}

Across tasks we evaluate baselines spanning three families: naive references, classical machine learning, and neural networks. The naive references are a global mean, a sector- or city-type group mean, and (company track only) an ExioML/EXIOBASE-derived emission-factor lookup mapping (GICS sector, ExioML region, year) to a tonnes-CO$_2$e estimate; forecasting tasks additionally use persistence, three-year moving average, linear trend, and a sector-growth extrapolation. Classical machine-learning baselines are Ridge regression, RandomForest~\citep{breiman2001random}, XGBoost~\citep{chen2016xgboost}, and LightGBM~\citep{ke2017lightgbm}, with a lag-feature gradient-boosted regressor reused for forecasting. Neural baselines comprise a multi-layer perceptron and a panel of pretrained foundation models: TabPFN~v2~\citep{hollmann2025tabpfn} for tabular regression; Chronos~\citep{ansari2024chronos}, TimesFM~\citep{das2024timesfm}, and Moirai~\citep{woo2024moirai} for time-series forecasting; and an LLM panel of Claude~\citep{anthropic2024claude3}, GPT~\citep{openai2023gpt4}, Qwen~\citep{yang2024qwen25}, and Mistral~\citep{jiang2023mistral} on the company track.

\section{Experiments and Analysis}
\label{sec:experiments}

We report headline results on both tracks under the protocol in \S\ref{sec:benchmark_stats}, organised around three benchmark-level findings: a cross-entity difficulty asymmetry (\S\ref{sec:exp_t1_difficulty}), an ID-to-OOD gap that dominates within-model variation (\S\ref{sec:exp_t2_cross_city}), and a multimodal lift concentrated on cross-city transfer (\S\ref{sec:exp_t2_s2}). Single-seed MLP numbers are flagged where they appear and are not used for ordering claims.

\subsection{Main Results}
\label{sec:exp_t1_structured}
\label{sec:exp_t1_difficulty}

\textbf{Company track.} Company-track performance is governed primarily by information regime, not by small differences between strong models. \textbf{(i)} With \emph{Open} features alone the eight learned baselines cluster at $R^2\!\approx\!0.28$--$0.31$ (Table~\ref{tab:t1_main}), barely above the sector-mean floor ($0.232$ on Open, $0.147$ on the matched panel)---country, year, and the ExioML factor add little beyond sector. \textbf{(ii)} The yfinance-matched subset drops $R^2$ by $0.03$ (TabPFN~v2) to $0.11$ (XGBoost), confirming the matched panel is genuinely harder; TabPFN~v2 leads on Open matched ($0.268$), consistent with its low-data prior advantage. \textbf{(iii)} Adding firm-level features lifts the strongest tree and TabPFN models to $R^2\!\approx\!0.86$--$0.88$---a $+0.59$ to $+0.70$ jump, an order of magnitude larger than the within-tree gap ($\le 0.008$); SICS sub-sector encoding adds a further $\sim\!+0.01$ on +firm, a smaller but consistent second-tier lever. On +firm TabPFN~v2 is paired-bootstrap-tied with every tuned tree ($|\Delta R^2|\!\leq\!0.017$, $p\!\geq\!0.14$), with trees partly distinguishable on shared rows (Appendix~\ref{app:paired}). \textbf{(iv)} Scope~3 follows the same regime pattern at a lower absolute level (best $R^2\!=\!0.688$, LightGBM+SICS); cross-region transfer is the only company-track setting where Ridge ($0.506\!\pm\!0.053$) beats the tuned trees (cross-region means $0.452$--$0.485$), consistent with linear models being less prone to overfit region-specific interactions. Auxiliary forecasting and LLM panels (Appendix Tables~\ref{tab:t1_scope3_forecasting},~\ref{tab:t1_llm_panel}) show simple temporal priors dominate annual forecasting and LLMs trail strong models by $0.32$--$1.12$~$R^2$ on the $n\!=\!200$ overlap.

\begin{table*}[t]
\centering
\caption{Company-track regression, $R^2$ (computed in $\log_{10}$ space; feature regimes defined in \S\ref{sec:dataset_t1}). Rows are grouped by method family: naive (Sector mean), classical ML (Ridge, tree ensembles), and deep / foundation (MLP, FT-Transformer, TabPFN~v2). \textbf{Cross-region}: leave-one-region-out over US/EU/APAC, mean on +firm. Bold: column max.}
\label{tab:t1_main}
\scriptsize
\setlength{\tabcolsep}{3.0pt}
\renewcommand{\arraystretch}{1.25}
\resizebox{\textwidth}{!}{%
\begin{tabular}{lccccccccc}
\toprule
\rowcolor{tabheader}
\multirow{2}{*}{\textbf{Model}} & \multicolumn{4}{c}{\cellcolor{tabheader}\textbf{Scope~1+2}} &
  \multicolumn{4}{c}{\cellcolor{tabheader}\textbf{Scope~3}} &
  \multicolumn{1}{c}{\cellcolor{tabheader}\textbf{Cross-region}} \\
\cmidrule(lr){2-5}\cmidrule(lr){6-9}\cmidrule(lr){10-10}
\rowcolor{tabheader}
& \textbf{Open} & \textbf{Open matched} & \textbf{+firm} & \textbf{+SICS} & \textbf{Open} & \textbf{Open matched} & \textbf{+firm} & \textbf{+SICS} & \textbf{+firm} \\
\midrule
\textbf{Sector mean} & 0.232 & 0.147 & 0.147 & 0.147 & 0.162 & 0.064 & 0.064 & 0.064 & 0.055 \\
\midrule
\rowcolor{tabrowalt}
\textbf{Ridge} & 0.284 & 0.229 & 0.549 & 0.613 & 0.211 & 0.128 & 0.370 & 0.411 & \textbf{0.506} \\
\textbf{RandomForest} & 0.288 & 0.201 & 0.875 & \textbf{0.885} & 0.181 & 0.059 & 0.639 & 0.687 & 0.485 \\
\rowcolor{tabrowalt}
\textbf{LightGBM} & 0.278 & 0.180 & 0.873 & 0.880 & 0.190 & 0.048 & 0.637 & \textbf{0.688} & 0.478 \\
\textbf{XGBoost} & 0.282 & 0.177 & \textbf{0.879} & \textbf{0.885} & 0.213 & 0.072 & 0.645 & 0.681 & 0.468 \\
\rowcolor{tabrowalt}
\textbf{HistGradientBoosting} & 0.300 & 0.235 & 0.871 & 0.869 & 0.215 & 0.059 & \textbf{0.653} & 0.677 & 0.452 \\
\midrule
\textbf{MLP} & 0.312 & 0.265 & 0.693 & 0.747 & 0.231 & 0.127 & 0.460 & 0.497 & 0.414 \\
\rowcolor{tabrowalt}
\textbf{FT-Transformer} & \textbf{0.313} & 0.246 & 0.622 & 0.690 & 0.212 & 0.129 & 0.418 & 0.461 & 0.494 \\
\textbf{TabPFN~v2} & 0.297 & \textbf{0.268} & 0.857 & 0.873 & \textbf{0.235} & \textbf{0.139} & 0.633 & 0.631 & 0.467 \\
\bottomrule
\end{tabular}
}
\end{table*}

\textbf{Building track.} Table~\ref{tab:t2_main} reports the core building regression setting. \textbf{(i)} Building prediction is intrinsically harder than company prediction: the strongest building model (TabPFN~v2 at $R^2\!=\!0.479$ on building-grouped) sits well below the strongest company model ($R^2\!=\!0.879$, Table~\ref{tab:t1_main}). The two tracks use different target scales, but the gap reflects a genuine task-level difference---company emissions are dominated by firm size (revenue alone is the top single predictor, Appendix~\ref{app:t1_feature_attribution}), while annual building emissions also depend on occupancy schedules, runtime, equipment efficiency, and occupant behaviour, factors that \emph{no public disclosure captures} and vary widely between similar buildings; the benchmark deliberately spans this heterogeneity (26 metros, three countries, 13 portals), and the lower attainable $R^2$ is a property of the task. \textbf{(ii) On unseen buildings, TabPFN~v2 is, to our knowledge, the first baseline to open a paired-bootstrap-significant gap over tuned trees on a multi-city, multi-source building-emissions panel.} Under the building-grouped split (every test building held out from train---the deployable setting), TabPFN~v2 reaches $R^2\!=\!0.482\!\pm\!0.029$, beating LightGBM by $+0.025$ ($p\!=\!.012$) and XGBoost by $+0.036$ ($p\!=\!.002$) on shared test rows, and tying with RandomForest ($+0.010$, $p\!=\!.36$); adding the NASA POWER climate channels makes the gap significant against \emph{all three} trees ($p\!\le\!.001$, Appendix~\ref{app:paired}). The ranking inverts under the row-random split, where RF leads ($0.477$ vs.\ TabPFN $0.405$)---confirming that the gap is specific to the unseen-entity regime and not an artefact of any tabular property of the data. \textbf{(iii)} Switching to building-grouped drops RF by $\sim\!6$ pp ($0.477\!\to\!0.415$) and trims tabular MLP modestly ($0.353\!\pm\!0.08\!\to\!0.341\!\pm\!0.08$); adding the four NASA POWER climate channels yields a similar grouped MLP at $0.346\!\pm\!0.06$. Tree splits on raw lat/lon already partition the space along climate-correlated boundaries, whereas MLP's dense first layer must learn this geographic prior end-to-end and trails the tuned trees by $\sim\!0.07$~$R^2$ on the grouped split. FT-Transformer, despite its richer transformer architecture, also reaches only $R^2\!=\!0.403\!\pm\!0.023$ on Core grouped, sitting between MLP and the tuned trees and never matching TabPFN. This confirms that the foundation-model edge comes from in-context pretraining rather than from transformer architecture alone.

Multimodal S2 results, cross-tier and foundation-model rankings, and analyses of schema and transfer effects are discussed in Section~\ref{sec:exp_t1_sector_factor} onwards; temporal hold-out (Appendix~\ref{app:task_b_grid}), cross-property-type stress (Appendix~\ref{app:task_d}), and forecasting (Appendix~\ref{app:t2_transfer_forecasting}) are reported as appendix-level evidence.

\begin{table*}[t]
\centering
\caption{Building-track regression on the 26-city cross-country core panel (\S\ref{sec:dataset_t2}). \textbf{+S2}: trees/TabPFN use concat fusion; MLP uses residual fusion (\S\ref{sec:exp_t2_s2}). Rows grouped by method family: naive, classical ML, deep / foundation. Bold: column max.}
\label{tab:t2_main}
\footnotesize
\setlength{\tabcolsep}{3.2pt}
\renewcommand{\arraystretch}{1.25}
\begin{tabular}{l ccc ccc}
\toprule
\rowcolor{tabheader}
\multirow{2}{*}{\textbf{Model}} & \multicolumn{3}{c}{\cellcolor{tabheader}\textbf{Row-random}} & \multicolumn{3}{c}{\cellcolor{tabheader}\textbf{Building-grouped}} \\
\cmidrule(lr){2-4} \cmidrule(lr){5-7}
\rowcolor{tabheader}
 & \textbf{Core} & \textbf{+ Climate} & \textbf{+ S2} & \textbf{Core} & \textbf{+ Climate} & \textbf{+ S2} \\
\midrule
\textbf{GlobalMean}   & $-0.045 \pm 0.002$ & $-0.045 \pm 0.002$ & --- & $-0.045 \pm 0.004$ & $-0.045 \pm 0.004$ & --- \\
\rowcolor{tabrowalt}
\textbf{CityTypeMean} & $\phantom{-}0.031 \pm 0.002$ & $\phantom{-}0.031 \pm 0.002$ & --- & $\phantom{-}0.030 \pm 0.005$ & $\phantom{-}0.030 \pm 0.005$ & --- \\
\midrule
\textbf{Ridge}        & $-3.47 \pm 1.81$ & $-3.36 \pm 1.83$ & $-3.98 \pm 2.11$ & $-3.86 \pm 1.47$ & $-3.75 \pm 1.41$ & $-3.08 \pm 2.01$ \\
\rowcolor{tabrowalt}
\textbf{RandomForest} & $\mathbf{0.477 \pm 0.030}$ & $\mathbf{0.476 \pm 0.028}$ & $\mathbf{0.444 \pm 0.012}$ & $\phantom{-}0.415 \pm 0.030$ & $\phantom{-}0.413 \pm 0.028$ & $\mathbf{0.427 \pm 0.021}$ \\
\textbf{LightGBM}     & $\phantom{-}0.413 \pm 0.020$ & $\phantom{-}0.414 \pm 0.020$ & $\phantom{-}0.401 \pm 0.017$ & $\phantom{-}0.401 \pm 0.023$ & $\phantom{-}0.401 \pm 0.026$ & $\phantom{-}0.399 \pm 0.012$ \\
\rowcolor{tabrowalt}
\textbf{XGBoost}      & $\phantom{-}0.410 \pm 0.020$ & $\phantom{-}0.415 \pm 0.023$ & $\phantom{-}0.401 \pm 0.017$ & $\phantom{-}0.399 \pm 0.025$ & $\phantom{-}0.399 \pm 0.026$ & $\phantom{-}0.397 \pm 0.013$ \\
\midrule
\textbf{MLP}          & $\phantom{-}0.353 \pm 0.076$ & $\phantom{-}0.343 \pm 0.080$ & $\phantom{-}0.334 \pm 0.150$ & $\phantom{-}0.341 \pm 0.078$ & $\phantom{-}0.346 \pm 0.059$ & $\phantom{-}0.385 \pm 0.078$ \\
\rowcolor{tabrowalt}
\textbf{FT-Transformer} & $\phantom{-}0.417 \pm 0.039$ & $\phantom{-}0.394 \pm 0.018$ & $\phantom{-}0.408 \pm 0.017$ & $\phantom{-}0.403 \pm 0.023$ & $\phantom{-}0.398 \pm 0.023$ & $\phantom{-}0.414 \pm 0.013$ \\
\textbf{TabPFN v2}    & $\phantom{-}0.405 \pm 0.071$ & $\phantom{-}0.409 \pm 0.067$ & $\phantom{-}0.377 \pm 0.065$ & $\mathbf{0.482 \pm 0.029}$ & $\mathbf{0.482 \pm 0.027}$ & $\phantom{-}0.380 \pm 0.060$ \\
\bottomrule
\end{tabular}
\vspace{-1.2em}
\end{table*}

\begin{figure}[t]
\centering
\includegraphics[width=\linewidth]{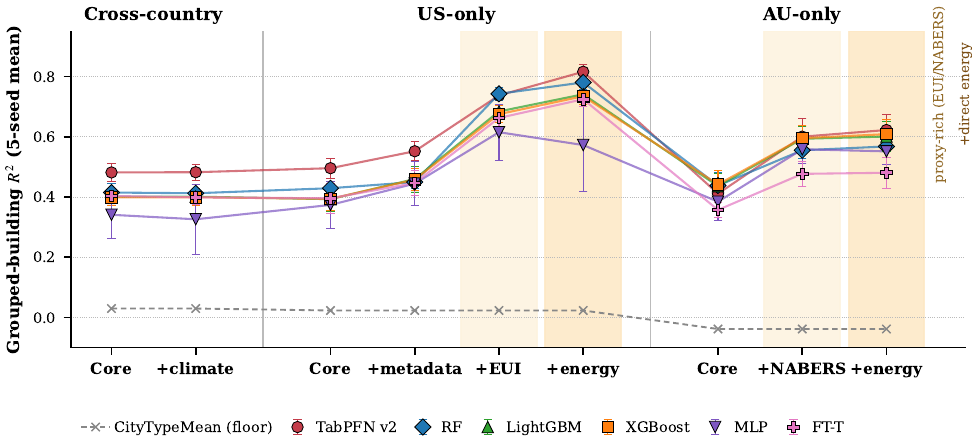}
\caption{Building-track $R^2$ on the building-grouped split across the nine feature tiers and three ladders defined in \S\ref{sec:feature_tiers} (full registry in Appendix~\ref{app:feature_tiers}). Shaded bands mark proxy-rich and direct-energy-proxy tiers.}
\label{fig:t2_tier_ladder_main}
\end{figure}

\subsection{Analysis and Findings}
\label{sec:exp_t1_sector_factor}

\textbf{Sector-factor estimation trails learned models.} Predicting
emissions by multiplying revenue with the ExioML/EXIOBASE sectoral
factor reaches $R^2=0.222$ on the firm-matched company panel, well below the tuned
tree family ($\approx\!0.87$); a sector-mean predictor sits even lower
at $0.147$. Sector identity is still informative when used as one
feature among many---adding SICS sub-sector target encoding lifts most
baselines (Table~\ref{tab:t1_main}), and tuned-LightGBM attribution
ranks the ExioML factor second after revenue---but it is too coarse to
drive prediction on its own. Per-feature attribution and a Financial
Services case study with a SICS-granularity check are in
Appendix~\ref{app:t1_feature_attribution} and
Appendix~\ref{app:t1_sector_case}.

\textbf{A pretrained tabular foundation model is consistently competitive and often best.} On the building track, a 1{,}000-resample paired bootstrap on shared test rows (Appendix~\ref{app:paired}, Table~\ref{tab:app_t2_paired}) shows TabPFN~v2 \emph{paired-bootstrap-significantly outperforms every tuned tree ($p\!<\!.05$) on three of six building-grouped tiers}---26-city core+climate, U.S. core, and U.S. metadata. On the 26-city core tier, TabPFN beats LightGBM ($p\!=\!.012$) and XGBoost ($p\!=\!.002$) but ties RF ($p\!=\!.36$); on the small Australia-only panel all four models are tied. On the company track TabPFN is also tied with the tuned trees ($R^2\!=\!0.857$, $p\!\geq\!0.14$). MLP, FT-Transformer, and LLM baselines remain weaker, so the advantage is specific to TabPFN's in-context training rather than a blanket win for deep tabular models~\citep{grinsztajn2022tabular}.

\label{sec:exp_t2_feature}
\label{sec:exp_t2_ranking}

\textbf{Feature quality drives rankings more than model choice.} Tuned-LightGBM permutation attribution on the 26-city core tier (Appendix~\ref{tab:app_t2_feature_attribution}) ranks gross floor area first, location second, and climate/time fields third; in Figure~\ref{fig:t2_tier_ladder_main}, moving from clean physical attributes to EUI/rating/direct-energy proxies changes the attainable $R^2$ far more than swapping among strong tree variants.

\label{sec:exp_t2_cross_city}
\label{sec:exp_t2_temporal}

\begin{figure}[t]
\centering
\includegraphics[width=\linewidth]{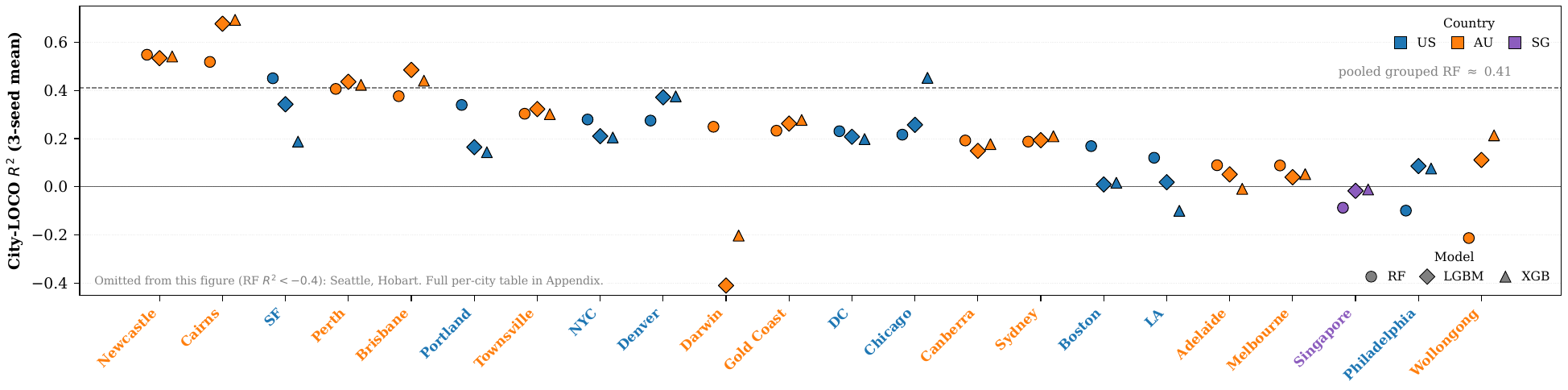}
\caption{Building-track leave-one-city-out on the 26-city cross-country core tier. Cities sorted by RF $R^2$ descending; dashed line is the pooled-panel Building-grouped RF baseline. Off-scale tree points (Seattle, Hobart) are flagged at the left edge. Per-city table for all models in Appendix~\ref{tab:app_task_c1_city_r2}.}
\label{fig:t2_cross_city}
\vspace{-0.8em}
\end{figure}

\textbf{Cross-city transfer is the hardest core building setting.}
Figure~\ref{fig:t2_cross_city} reports three-seed leave-one-city-out results
on the 26-city cross-country core tier. Mean-city tree-family $R^2$
collapses from the pooled-panel Building-grouped value of $\sim\!0.41$ (Table~\ref{tab:t2_main}) to
$R^2\!=\!0.03$ (LightGBM), $0.08$ (XGBoost), and $0.13$ (RandomForest). The pattern is
heterogeneous rather than a uniform failure: Newcastle, Cairns, and
Perth remain at $R^2\!\approx\!0.40$--$0.69$ across all three tuned trees, while Seattle
and Hobart turn strongly negative (RF $-1.23$
on Seattle; LGBM/XGB collapse to below $-2$ on Hobart). The three tree
families largely agree on which cities are easy versus hard, indicating
the failures are driven by city-level distribution shift rather than a
single model's brittleness. The full per-city table for all models,
including the off-scale tabular MLP failures, is in
Appendix~Table~\ref{tab:app_task_c1_city_r2}.

\label{sec:exp_t2_s2}

\textbf{Sentinel-2 helps cross-city transfer, not in-distribution prediction.} On the S2-eligible subset, PCA-64 Sentinel-2~+~Clay embeddings are nearly a no-op in-distribution ($\Delta R^2\!\in\![-0.012,-0.007]$ for trees on the grouped split) because trees already capture geography through lat/lon. On cross-city leave-one-out, mean-city $R^2$ rises by $+0.071$ (LightGBM, $+0.054\!\to\!+0.125$) and $+0.038$ (XGBoost), with 11/23 and 14/23 city win-rates respectively; raw lat/lon are unique tokens that fail to transfer, while a Sentinel-2 patch's visual content (rooftop type, density, vegetation) supplies a generalisable geographic prior. We position Sentinel-2~+~Clay embeddings as a transfer-augmenting signal, not a uniform improver---useful precisely where tabular generalisation breaks.

\textbf{The cross-city S2 effect is heterogeneous.} San Francisco, Washington DC, and Melbourne gain on all three models, with Boston and Perth additionally gaining on both tree models, while Chicago loses on both trees. S2 is strongest where dense urban form yields a transferable visual prior, weakest when the training panel is dominated by similar-schema cities. A fusion-design ablation and per-city deltas are in Appendix~\ref{app:s2_supp}.

\textbf{Forecasting reveals a panel-length asymmetry across tracks.} On the company track's six-year panel, naive temporal references (sector-growth and linear trend) reach $R^2\!\approx\!0.94$--$0.96$, while TimesFM and Chronos trail at $0.81/0.78$ (Appendix~Table~\ref{tab:t1_scope3_forecasting}). On the building track's 5-year panel, time-series foundation models lead at every horizon $h\!\in\!\{2,\ldots,5\}$ (TimesFM peaks at $R^2\!=\!0.90$ on $h\!=\!4$), with tuned trees within $\sim\!0.1$~$R^2$ and naive linear trend lagging by $\sim\!0.2$ (Appendix~Table~\ref{tab:app_task_e_horizon}). Building forecasting therefore rewards larger temporal models, while the shorter company panel leaves simple temporal priors hard to beat.

\section{Conclusion}
\label{sec:conclusion}

\textbf{GHGbench}'s primary contribution is a harmonised cross-country GHG dataset with canonical in-distribution and cross-region/city splits and a paired-bootstrap protocol. Three findings emerge: a cross-entity difficulty asymmetry, an ID-to-OOD gap that dominates within-model variation, and a multimodal lift concentrated where tabular generalisation breaks. Longer-horizon forecasting, frontier LLMs, and richer multimodal fusion are open follow-ups the benchmark is designed to support.

\section*{Acknowledgments}
This research is supported by the ARC Training Centre for Whole Life Design of
Carbon Neutral Infrastructure (IC230100015).

%%%%%%%%%%%%%%%%%%%%%%%%%%%%%%%%%%%%%%%%%%%%%%%%%%%%%%%%%%%%
\bibliographystyle{plainnat}
\bibliography{refs}

%%%%%%%%%%%%%%%%%%%%%%%%%%%%%%%%%%%%%%%%%%%%%%%%%%%%%%%%%%%%
\clearpage
\appendix
\section{Datasheet for GHGbench}
\label{app:datasheet}

\textbf{Motivation.} GHGbench supports machine-learning research on greenhouse-gas (GHG) emission
prediction for two complementary entity types: companies and buildings. The
two tracks were assembled because no public benchmark previously evaluated
deep, foundation, and multimodal models on heterogeneous, multi-source
emissions data under realistic generalisation regimes.

\textbf{Composition.} The company track contains 32{,}830 enriched company--year rows from 12{,}087 companies
(2018--2023), with 31{,}331 usable Scope~1+2 labels and 18{,}763 lenient-coverage
Scope~3 labels after the filters described in
Section~\ref{sec:dataset_t1}. The building track contains 491{,}591 building--year rows
across 100{,}984 buildings and 26 metropolitan areas in the United States,
Australia, and Singapore, of which 471{,}070 carry non-null annual GHG
labels. Each row is augmented with NASA POWER climate features and
optionally with Sentinel-2 + Clay image embeddings (369{,}698 rows have a
valid building-footprint patch in the headline multimodal subset).

\textbf{Collection process.} Company labels and disclosures come from CDU exports; financial enrichment
is built from yfinance and matched to a ticker. Building disclosures come
from 13 city- or country-level open-data portals; raw fields are normalised
through the harmonisation scripts described in
Section~\ref{sec:dataset_t2}, including a NABERS$\to$kBTU unit fix and a
property-type taxonomy of 22 categories.

\textbf{Recommended uses.} The benchmark is designed for in-distribution regression, temporal hold-out,
cross-region and cross-city transfer, cross-property-type stress testing,
short-horizon forecasting, and multimodal alignment. We discourage using the
proxy-rich and direct-energy-proxy tiers as deployable headline results;
Section~\ref{sec:feature_tiers} formalises the reporting rule.

\textbf{Maintenance.} Source data are not redistributed in raw form. The release is a
reconstruction recipe: code, canonical splits, harmonisation metadata, and
download instructions for each portal. Updates will follow source-data
releases (annual disclosure cycles) and will be versioned.

Table~\ref{tab:dataset_overview} summarises the two tracks in compact form.

\begin{table}[t]
\centering
\caption{GHGbench dataset overview.}
\label{tab:dataset_overview}
\footnotesize
\resizebox{\linewidth}{!}{%
\begin{tabular}{p{0.15\linewidth} p{0.32\linewidth} p{0.20\linewidth} p{0.35\linewidth}}
\toprule
Track & Scale & Targets & Key signals \\
\midrule
Company
& 12{,}087 companies; 32{,}830 enriched company--year rows from 2018--2023; 31{,}331 usable Scope~1+2 rows; 18{,}763 filtered Scope~3 rows.
& Annual Scope~1+2 and Scope~3 emissions.
& Country, sector, financial features, ExioML/EXIOBASE-derived sectoral factors, and disclosure text/business summaries.
\\
\midrule
Building
& 100{,}984 buildings; 491{,}591 building--year rows from 13 open sources across 26 metros in the United States, Australia, and Singapore; reporting years 2011--2026, dense through 2024.
& Annual building operating GHG emissions in metric tonnes CO$_2$e.
& Harmonised building attributes, property taxonomy, coordinates, NASA POWER building-year climate, and Sentinel-2/Clay image embeddings.
\\
\bottomrule
\end{tabular}%
}
\end{table}

\section{Limitations and Broader Impact}
\label{app:limitations}

\textbf{Limitations.} GHGbench's headline contribution is the harmonised dataset, canonical splits, and paired-bootstrap protocol on in-distribution and cross-region/city transfer; forecasting, LLM-only emission prediction, and advanced multimodal fusion are reported as supplementary axes rather than headline claims, and the limitations below are scoped accordingly. The public-data design also sets limits. The company track depends on CDU self-reported
disclosures and is released as a reconstruction recipe rather than a raw-data
mirror. The building track covers 26 metropolitan areas and 13 sources, but coverage is skewed
toward cities that publish benchmarking data; some cities are temporally sparse,
and Los Angeles and Denver rely heavily on address-level geocoding. The
Sentinel-2 extension uses only valid building-footprint patches in headline
multimodal experiments, and the model panel is baseline-oriented rather than an
exhaustive architecture search. Annual forecasting is evaluated only one step
ahead on a short five-year panel, so longer-horizon time-series foundation-model
behaviour is undersampled. Multimodal fusion uses only late concatenation of
PCA-compressed Sentinel-2 embeddings, and LLM-only emission prediction is
evaluated as zero/few-shot point regression; we leave more advanced multimodal
fusion architectures and reasoning-style LLM pipelines as open follow-up work.

\textbf{Broader impact.} GHGbench is intended to support open, comparable evaluation of carbon-emission prediction for climate-policy and urban-operations research that previously depended on paywalled disclosure feeds. As with any predictive benchmark, point-estimate predictions can convey false confidence on noisy panels; we encourage users to consult the per-cell standard deviations and bootstrap confidence intervals we report rather than headline means, especially on small or out-of-distribution panels (e.g., the AU NABERS panel) where seed variance is high.

\section{Data Harmonisation Details}
\label{app:data_harmonisation}

\textbf{Company preprocessing.} The company track converts dash placeholders and disclosure fields to numeric
tonnes CO$_2$e, constructs Scope~1+2 from reported Scope~1 and
location-based Scope~2, normalises company identifiers and country codes, and
deduplicates company--year records before splitting. For Scope~1+2, labels in
$[0,10)$ tonnes CO$_2$e or above $5\times10^8$ tonnes CO$_2$e are removed as
unit or reporting errors. Scope~3 is joined from the raw CDU disclosure file
at run time and filtered to records with non-null country and Scope~3 between
100 and $2\times10^9$ tonnes CO$_2$e. The lenient-coverage panel uses country,
reporting year, and sector metadata; the strict-coverage panel additionally requires a
ticker match and positive revenue, and adds time-aligned or fallback
financial variables.

\textbf{Building preprocessing.} The building track uses source-specific parsers for CSV, XLSX, and XLSB
portals; converts floor area to square feet, EUI to kBtu/ft$^2$, gas use to
kBtu, electricity to kWh, and GHG labels to metric tonnes CO$_2$e; and
collapses duplicate building--years using numeric medians and categorical
modes. Physically invalid values, including negative EUI/GHG values and
extreme EUI outliers, are removed or blanked while preserving rows with other
usable fields. The property-type mapping keeps coarse source labels explicit:
for example, San Francisco's raw ``Commercial'' label maps to a dedicated
mixed-commercial bucket rather than retail. The \texttt{coord\_source} column
separates source coordinates from address-level geocodes.

\section{Task-suite Details}
\label{app:tasks}

Table~\ref{tab:tasks} gives the full task definition table referenced in
Section~\ref{sec:benchmark_tasks}.

\begin{table*}[t]
\centering
\caption{GHGbench task suite.}
\label{tab:tasks}
\footnotesize
\resizebox{\textwidth}{!}{%
\begin{tabular}{p{0.08\textwidth} p{0.10\textwidth} p{0.16\textwidth} p{0.36\textwidth} p{0.22\textwidth}}
\toprule
Track & Task & Target & Train/test rule & Evaluation focus \\
\midrule
T1 & A & Scope~1+2 emissions & Stratified company--year split, 80/10/10 train/validation/test, stratified by GICS sector and country with rare-stratum fallback; reported on T1-Wide and T1-Strict. & In-distribution company regression and value of matched financial/text features. \\
T1 & A$'$ & Scope~3 emissions & Scope~3 is joined from raw CDU disclosures, filtered as in Section~\ref{sec:dataset_t1}, then rebuilt as an 80/10/10 stratified split with the same wide/strict definitions as T1-A. & Harder indirect-emissions target under the same company protocol. \\
T1 & C & Scope~1+2 emissions & Leave-one-region-out on T1-Strict: train on two of US, EU, APAC and evaluate on the held-out region. & Cross-region corporate generalization. \\
T1 & E & Scope~1+2 forecasts & Balanced 2018--2022 company panel; use 2018--2021 context to forecast 2022, with classical, ML, and time-series foundation-model baselines. & Short annual-panel forecasting and persistence-style baselines. \\
\midrule
T2 & A & Building-year GHG emissions & Pooled and per-city prediction with either row-random or grouped-building 70/10/20 train/validation/test splits; the grouped split is the main-paper deployable setting because each building appears in only one fold. & Feature-tier ladder, target-proxy interpretation, and baseline model ranking. \\
T2 & B & Building-year GHG emissions & Temporal hold-out: train on years $\leq$2019, validate on 2020, test on years $\geq$2021; COVID variant trains on years $\leq$2018 and evaluates 2019--2021 separately. & Temporal robustness under public-disclosure drift. \\
T2 & C1 & Building-year GHG emissions & Leave-one-city-out: train on all source cities, carve validation from source buildings, and test on the held-out city. & Cross-city transfer and country/schema heterogeneity. \\
T2 & D & Building-year GHG emissions & Leave-one-property-type-out on U.S. tiers with property-type metadata; main stress-test types are Office, Multifamily, Retail, Hotel, K-12 School, Hospital/Medical, and Warehouse/Distribution. & Cross-type stress test; appendix-level supporting evidence. \\
T2 & E & One-step building-year GHG forecasts & Eligible buildings must have enough history and data on both sides of the time cut; train on years $\leq$2019, validate on 2020, and test on years $\geq$2021 using lagged emissions and lagged climate only. & Short-horizon forecasting without current-year climate information. \\
T2 & S2 & Building-year GHG emissions & Re-run T2-A and T2-C1 on the identical S2-eligible subset for tabular-only and tabular-plus-Sentinel variants; S2 uses raw 1024-dimensional embeddings or PCA-64/PCA-128 fitted on train embeddings only. & Multimodal extension under matched row availability. \\
\bottomrule
\end{tabular}%
}
\end{table*}

\section{Splits and Generalization Axes}
\label{app:splits}

This section gives the concrete split parameters for the four evaluation
capabilities defined in Section~\ref{sec:benchmark_tasks}.

\textbf{In-distribution splits.} Company regression uses a deterministic 80/10/10 train/validation/test split stratified over (GICS sector, country) with rare-stratum fallback, reported on both the lenient-coverage and strict-coverage panels defined in Section~\ref{sec:dataset_t1}. The lenient-coverage panel illustrates how far broad-coverage metadata alone can go, while the strict-coverage panel carries the headline structured, LLM, and paired-bootstrap comparisons because those require matched financial and text features. The Scope~3 task applies the row filter from Section~\ref{sec:dataset_t1} and \emph{rebuilds} an 80/10/10 stratified split on the filtered panel rather than reusing the Scope~1+2 split. Building regression uses a 70/10/20 split in two variants: a row-random split for sensitivity analyses, and a grouped-building split that keeps every building in a single fold. \emph{The grouped variant is the main-paper deployable setting} because evaluation buildings are unseen during training.

\textbf{Cross-distribution splits.} Company cross-region transfer leaves one of \{U.S., EU, APAC\} out at a time and is evaluated on the strict-coverage panel. Building cross-city transfer holds out one entire metro at a time, splitting source-city buildings between training and validation while the held-out city contributes only test rows. Building cross-property-type transfer holds out one of seven property categories at a time, reported as appendix-level evidence.

\textbf{Temporal split.} The building-level temporal hold-out trains on years $\le 2019$, validates on 2020, and tests on years $\ge 2021$; Denver, Philadelphia, and Portland have only one or three reporting years (Section~\ref{sec:dataset_t2}) and are excluded. The company panel covers only 2018--2023, so company-level temporal generalization is evaluated through forecasting rather than a separate hold-out.

\textbf{Forecasting splits.} Company forecasting uses a balanced 2018--2022 panel and predicts 2022 emissions from 2018--2021 context. Building forecasting uses the same year cuts as the temporal hold-out but restricts inputs to lagged emissions and lagged climate, preventing current-year climate leakage.

\section{Property-type Taxonomy}
\label{app:property_taxonomy}

Table~\ref{tab:app_property_taxonomy} lists the 22 canonical building
property-type categories used in the building track, the typical raw forms each category
absorbs, and the cities for which the category is populated. The full mapping
JSON is available at \texttt{scripts/property\_type\_mapping.json}, and the
heuristic fallback used when raw values do not match the mapping is in
\texttt{scripts/standardize\_buildings.py}.

\begin{table}[t]
\centering
\caption{T2 canonical property-type taxonomy (22 categories). Mapping: \texttt{scripts/property\_type\_mapping.json}.}
\label{tab:app_property_taxonomy}
\footnotesize
\resizebox{0.92\linewidth}{!}{%
\begin{tabular}{p{0.20\linewidth} p{0.35\linewidth} p{0.29\linewidth}}
\toprule
Category & Typical raw forms mapped here & Cities where populated \\
\midrule
\texttt{Office} & ''Office'', ''Medical Office'', ''Bank Branch'' & nyc, la, seattle, dc, chicago, boston, portland \\
\texttt{Multifamily Housing} & ''Multifamily Housing'', ''Multifamily'', ''Residential'', ''Mixed Residential'', ''Other - Lodging/Residential'' & all US + singapore \\
\texttt{Retail} & ''Retail Store'', ''Mall'', ''Shopping Center'' (NOT bare ''Commercial'') & nyc, la, seattle, dc, chicago \\
\texttt{Hotel} & ''Hotel'', ''Motel'', ''Lodging'' & all US \\
\texttt{K-12 School} & ''K-12 School'', ''School'' & nyc, dc, chicago, seattle \\
\texttt{College/University} & ''College/University'', ''University'' & nyc, dc, chicago \\
\texttt{Hospital/Medical} & ''Hospital'', ''Medical Office'', ''Clinic'' & all US \\
\texttt{Warehouse/Distribution} & ''Warehouse'', ''Distribution Center'' & nyc, la, seattle, dc \\
\texttt{Industrial} & ''Manufacturing/Industrial'', ''Industrial'' & nyc, la \\
\texttt{Worship} & ''Worship Facility'', ''Church'', ''Synagogue'', ''Mosque'' & nyc, la, seattle, dc \\
\texttt{Senior Living} & ''Senior Living Community'', ''Nursing Home'' & most US \\
\texttt{Supermarket/Grocery} & ''Supermarket/Grocery Store'' & la, chicago, seattle \\
\texttt{Restaurant} & ''Restaurant'', ''Food Service'' & small \\
\texttt{Parking} & ''Parking'', ''Garage'' & la, dc \\
\texttt{Laboratory} & ''Laboratory'' & seattle, chicago \\
\texttt{Fitness Center} & ''Fitness Center/Health Club'' & small \\
\texttt{Library} & ''Library'' & dc \\
\texttt{Residence Hall} & ''Residence Hall/Dormitory'' & nyc, dc \\
\texttt{Mixed Use} & ''Mixed Use Property'', ''RES/COMMERCIAL USE'' & nyc, la, dc \\
\texttt{Self-Storage} & ''Self-Storage Facility'' & nyc, la \\
\texttt{Commercial (Mixed)} & ''Commercial'', ''Commercial - Port Facility'' & sf only -- SF raw data uses one coarse ''Commercial'' bucket that covers offices, retail, and mixed-use; we keep it as a separate category to avoid inflating the Retail share. \\
\texttt{Other} & ''Other'', ''Mixed Use - Commercial'', residuals & all \\
\bottomrule
\end{tabular}
}
\end{table}

\section{Feature Availability across Cities and Tiers}
\label{app:feature_availability}

\begin{figure}[t]
  \centering
  \includegraphics[width=\linewidth]{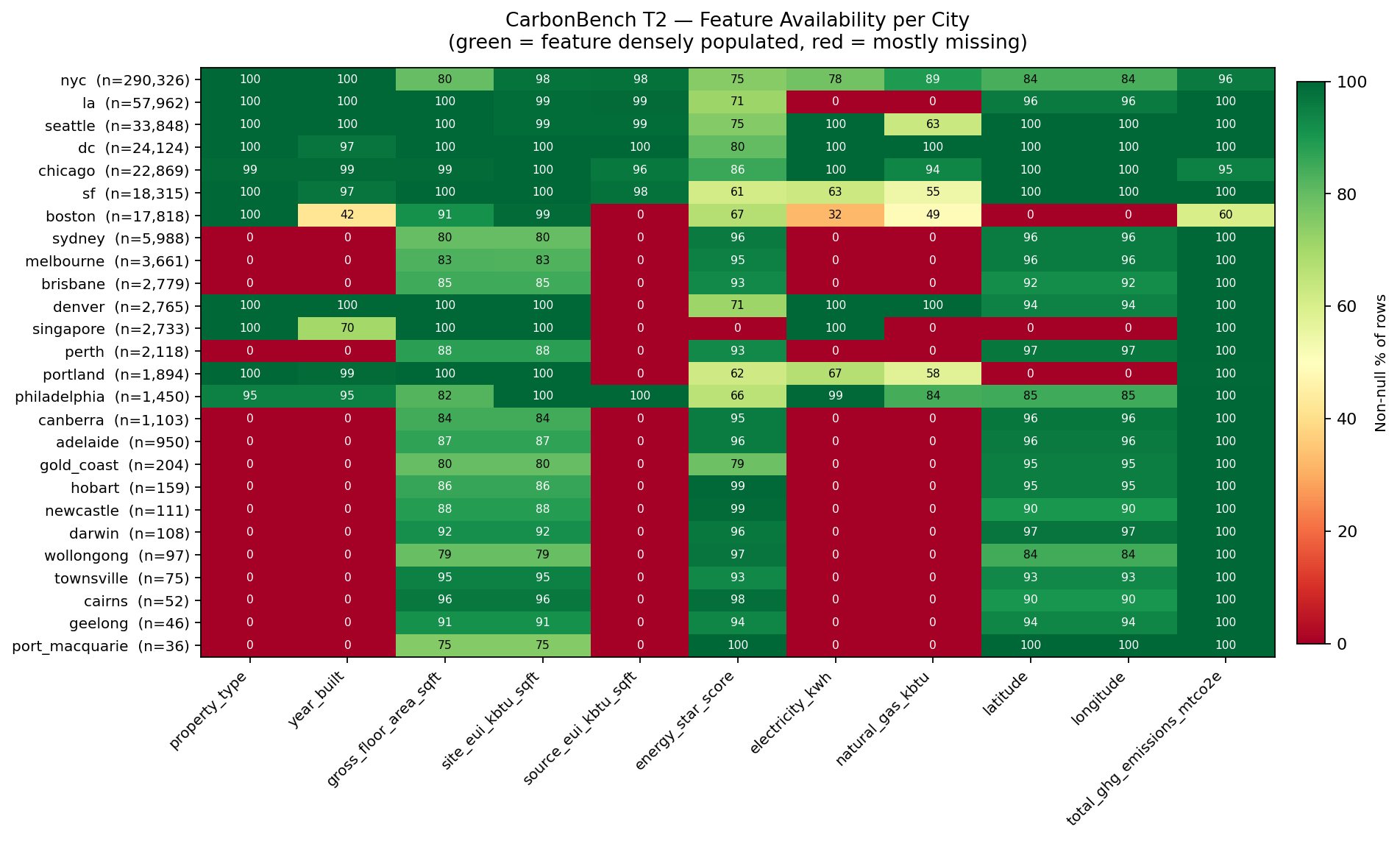}
  \caption{Per-city non-null availability (\%) for building-level schema
  fields. Cells at $0\%$ are hard exclusions for the corresponding tier.}
  \label{fig:app_feature_availability}
\end{figure}

Figure~\ref{fig:app_feature_availability} reproduces the per-city
non-null feature availability matrix, computed by
\texttt{scripts/feature\_availability.py}. The matrix is the design
constraint behind the feature-tier registry: cells with $0\%$ availability
on a given city are hard blockers and force separate per-country tiers
(notably AU has $0\%$ \texttt{property\_type} and \texttt{year\_built} in the
NABERS/BEEC schema, and Singapore has $0\%$ \texttt{site\_eui} but $100\%$
\texttt{source\_eui}).

\section{Building Feature-tier Registry}
\label{app:feature_tiers}

Table~\ref{tab:feature_tiers} lists the feature-tier registry used by the building track.
Clean tiers are the deployable settings; proxy-rich and direct-energy-proxy tiers
are retained only to quantify the value and risk of richer disclosure fields.

\begin{table}[t]
\centering
\caption{Building-track feature tiers. Clean tiers exclude energy-use proxies; proxy-rich tiers add EUI/rating; direct-energy-proxy tiers add raw electricity/gas. Proxy tiers are never pooled with clean tiers.}
\label{tab:feature_tiers}
\footnotesize
\resizebox{\linewidth}{!}{%
\begin{tabular}{@{}l p{0.08\linewidth} p{0.24\linewidth} p{0.12\linewidth} p{0.20\linewidth}@{}}
\toprule
Tier & Cities & Added features beyond size, coordinates, year, HDD, and CDD & Paper label & Role in paper \\
\midrule
\texttt{core\_all\_cities}
& 26
& None.
& Clean
& Main cross-country headline tier.
\\
\texttt{core\_all\_cities\_climate\_plus}
& 26
& Annual mean temperature, relative humidity, surface solar radiation, and wind speed from NASA POWER.
& Clean
& Climate ablation against the core tier.
\\
\texttt{us\_core}
& 6 U.S.
& Property type and year built.
& Clean
& U.S. metadata-rich baseline with fixed city population.
\\
\texttt{us\_metadata}
& 6 U.S.
& \texttt{us\_core} plus ENERGY STAR score.
& Clean
& Tests non-energy metadata value in U.S. disclosures.
\\
\texttt{us\_leaky\_eui}
& 6 U.S.
& \texttt{us\_metadata} plus site and source EUI.
& Proxy-rich
& Upper-reference tier; not a deployable headline result.
\\
\texttt{us\_leaky\_full}
& 6 U.S.
& \texttt{us\_leaky\_eui} plus electricity and natural-gas use.
& Direct-energy proxy
& Upper-reference tier with proximate target signals.
\\
\texttt{au\_core}
& 15 AU
& Climate-plus core features only.
& Clean
& Australia-only clean-schema comparison.
\\
\texttt{au\_eui}
& 15 AU
& \texttt{au\_core} plus site EUI.
& Proxy-rich
& Australia-only EUI reference tier.
\\
\texttt{au\_full}
& 15 AU
& \texttt{au\_eui} plus a NABERS-derived rating score.
& Proxy-rich
& Australia-only rating/EUI reference tier.
\\
\bottomrule
\end{tabular}%
}
\end{table}

\begin{figure}[t]
  \centering
  \includegraphics[width=\linewidth]{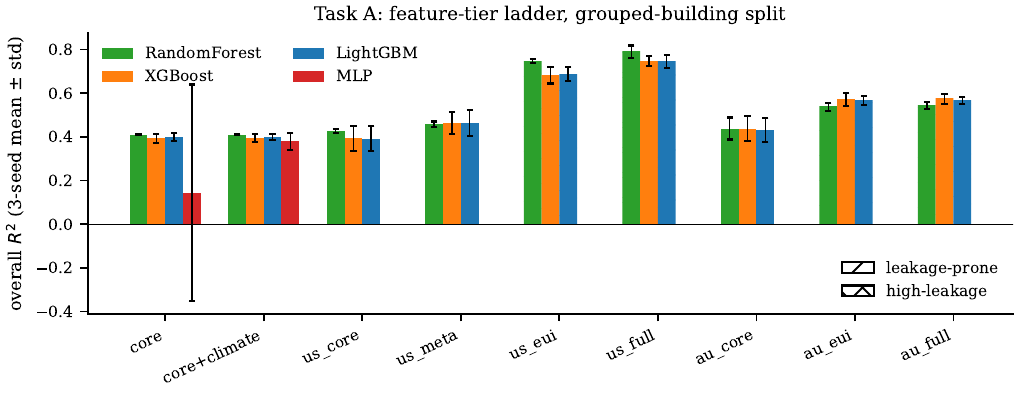}
  \caption{building-track regression feature-tier ladder, grouped-building split, 3-seed
  mean$\pm$std. Hatching: \texttt{//} proxy-rich, \texttt{xx}
  direct-energy-proxy. MLP init-only 5-seed refit:
  Table~\ref{tab:app_mlp_5seed}.}
  \label{fig:t2_feature_ladder}
\end{figure}

Table~\ref{tab:t2_tier_summary} reports the same tier ladder as a numerical
summary, complementing Figure~\ref{fig:t2_tier_ladder_main} in the main text.

\begin{table}[t]
\centering
\caption{Best point-estimate grouped-building $R^2$ on T2 Task~A across the nine feature tiers (best of tuned trees and TabPFN~v2 per tier).}
\label{tab:t2_tier_summary}
\footnotesize
\setlength{\tabcolsep}{4.5pt}
\begin{tabular}{llcc}
\toprule
Tier & Coverage / added signal & Best model & $R^2$ \\
\midrule
Cross-country core & 26 cities, clean physical core & TabPFN v2 & $0.479 \pm 0.024$ \\
Cross-country + weather & Core + NASA POWER weather & TabPFN v2 & $0.480 \pm 0.022$ \\
U.S. core & 6 U.S. cities, property/year metadata & TabPFN v2 & $0.511 \pm 0.019$ \\
U.S. metadata & U.S. core + ENERGY STAR score & TabPFN v2 & $0.563 \pm 0.032$ \\
U.S. EUI proxy & U.S. + EUI proxy fields & RandomForest & $0.747 \pm 0.010$ \\
U.S. direct-energy proxy & U.S. + electricity/gas proxies & TabPFN v2 & $0.821 \pm 0.028$ \\
AU core & 15 AU metros, clean AU schema & RandomForest & $0.438 \pm 0.050$ \\
AU EUI proxy & AU + EUI proxy field & TabPFN v2 & $0.591 \pm 0.083$ \\
AU rating/EUI proxy & AU + EUI and NABERS-derived rating & TabPFN v2 & $0.615 \pm 0.069$ \\
\bottomrule
\end{tabular}
\end{table}

\section{Statistical Protocol}
\label{app:protocol}

\textbf{Hyperparameter tuning.} Tree baselines are tuned with validation-based random search: 40 trials selecting by validation $R^2$ on the strict-coverage panel for the company track, and 15 trials selecting by validation log-MAE on the in-distribution building task. The MLP uses the same 15-trial budget over the search space listed in \texttt{scripts/run\_task\_a\_mlp\_hpsearch.py}. Best retained configurations per (feature tier, model) are listed in Appendix~\ref{app:hp_configs}.

\textbf{Multi-seed reporting.} Headline company Scope~1+2 baselines are reported with 5-seed mean and standard deviation. The building-track headline regression (Table~\ref{tab:t2_main}) and main-text feature-tier ladder (Figure~\ref{fig:t2_tier_ladder_main}) are 5-seed mean$\pm$std; the appendix per-tier feature ladder (Figure~\ref{fig:t2_feature_ladder}) and per-city cross-city transfer tables report 3-seed mean and standard deviation, since the 2-seed extension was applied only to the headline grouped-split panel. The MLP combines split-induced and initialisation-induced variance; an init-only 5-seed refit on a fixed split is reported in Appendix~\ref{app:mlp_5seed} (Table~\ref{tab:app_mlp_5seed}), and single-seed MLP numbers are not used for model-ordering claims.

\textbf{Uncertainty and significance testing.} Per-metric uncertainty is reported as a 1000-sample bootstrap confidence interval. Headline model-ordering claims---company structured-vs-LLM, tuned-tree-family internal comparisons, and building in-distribution model-ordering---are accompanied by a 1000-sample paired bootstrap on shared test rows. Full pair-wise $\Delta R^2$ values and $p$-values are tabulated in Appendix~\ref{app:paired}; we treat $p<0.05$ as the threshold for explicit ordering claims, and absence of paired-bootstrap evidence is reported as a tie rather than as a model-superiority claim.

\section{Hyperparameter Search and Best Configurations}
\label{app:hp_configs}

building-track regression tree baselines are tuned with 15-trial validation-based random
search; the validation metric is log-MAE on a held-out validation slice
created by the same grouped-building split as the test set. Building-track MLP baselines
use the same 15-trial budget over the search space listed in
\texttt{scripts/run\_task\_a\_mlp\_hpsearch.py}. Company-track tree baselines use a
40-trial random search selecting by validation $R^2$ on the strict-coverage panel.

Table~\ref{tab:app_hp_configs} reports the best configuration retained for
each (feature tier, model) on building-track regression, together with the test-set $R^2$ and
log-MAE produced by re-fitting that configuration. Differences within the
tuned tree family are small, consistent with the in-tier paired-bootstrap
results in Section~\ref{sec:exp_t2_ranking}.

{\tiny
\setlength{\tabcolsep}{3pt}
\begin{longtable}{>{\raggedright\arraybackslash}p{0.12\linewidth} l r r r >{\raggedright\arraybackslash}p{0.55\linewidth}}
\caption{T2 Task~A best HP configs (15-trial random search per (tier, model), selected by validation log-MAE).}
\label{tab:app_hp_configs} \\
\toprule
Feature tier & Model & Trials & Test $R^2$ & Test LogMAE & Best configuration \\
\midrule
\endfirsthead
\multicolumn{6}{c}{\tablename\ \thetable\ -- continued} \\
\toprule
Feature tier & Model & Trials & Test $R^2$ & Test LogMAE & Best configuration \\
\midrule
\endhead
\bottomrule
\endfoot
\texttt{au\_core} & LightGBM & 15 & 0.421 & 0.411 & n\_estimators=800, num\_leaves=31, learning\_rate=0.08, min\_child\_samples=20, reg\_alpha=1.0, reg\_lambda=0.1, colsample\_bytree=0.7, subsample=1.0, subsample\_freq=1 \\
\texttt{au\_core} & MLP & 15 & 0.392 & 0.409 & hidden\_layers=[512, 512, 256], dropout=0.1, lr=0.003, weight\_decay=0.0001, batch\_size=2048, max\_epochs=200, patience=20 \\
\texttt{au\_core} & RandomForest & 15 & 0.425 & 0.399 & n\_estimators=500, max\_depth=10, min\_samples\_leaf=5, max\_features=0.5 \\
\texttt{au\_core} & XGBoost & 15 & 0.410 & 0.418 & n\_estimators=800, max\_depth=8, learning\_rate=0.05, min\_child\_weight=3.0, subsample=1.0, colsample\_bytree=1.0, reg\_alpha=0.0, reg\_lambda=5.0 \\
\texttt{au\_eui} & LightGBM & 15 & 0.551 & 0.209 & n\_estimators=800, num\_leaves=255, learning\_rate=0.05, min\_child\_samples=5, reg\_alpha=0.1, reg\_lambda=5.0, colsample\_bytree=1.0, subsample=0.7, subsample\_freq=1 \\
\texttt{au\_eui} & MLP & 15 & 0.540 & 0.213 & hidden\_layers=[256, 128, 64], dropout=0.0, lr=0.001, weight\_decay=0.0001, batch\_size=1024, max\_epochs=200, patience=20 \\
\texttt{au\_eui} & RandomForest & 15 & 0.555 & 0.217 & n\_estimators=500, max\_depth=20, min\_samples\_leaf=2, max\_features=0.7 \\
\texttt{au\_eui} & XGBoost & 15 & 0.549 & 0.203 & n\_estimators=800, max\_depth=4, learning\_rate=0.08, min\_child\_weight=1.0, subsample=0.8, colsample\_bytree=0.9, reg\_alpha=1.0, reg\_lambda=5.0 \\
\texttt{au\_full} & LightGBM & 15 & 0.560 & 0.203 & n\_estimators=800, num\_leaves=127, learning\_rate=0.08, min\_child\_samples=20, reg\_alpha=1.0, reg\_lambda=0.1, colsample\_bytree=1.0, subsample=0.8, subsample\_freq=1 \\
\texttt{au\_full} & MLP & 15 & 0.513 & 0.228 & hidden\_layers=[256, 256, 128], dropout=0.1, lr=0.003, weight\_decay=1e-05, batch\_size=2048, max\_epochs=200, patience=20 \\
\texttt{au\_full} & RandomForest & 15 & 0.565 & 0.214 & n\_estimators=200, max\_depth=20, min\_samples\_leaf=1, max\_features=0.7 \\
\texttt{au\_full} & XGBoost & 15 & 0.555 & 0.204 & n\_estimators=800, max\_depth=8, learning\_rate=0.05, min\_child\_weight=10.0, subsample=1.0, colsample\_bytree=1.0, reg\_alpha=1.0, reg\_lambda=5.0 \\
\texttt{core} & LightGBM & 15 & 0.421 & 0.622 & n\_estimators=800, num\_leaves=127, learning\_rate=0.05, min\_child\_samples=5, reg\_alpha=1.0, reg\_lambda=0.1, colsample\_bytree=1.0, subsample=0.9, subsample\_freq=1 \\
\texttt{core} & MLP & 15 & -0.421 & 0.651 & hidden\_layers=[256, 256, 128], dropout=0.1, lr=0.003, weight\_decay=0.0001, batch\_size=4096, max\_epochs=200, patience=20 \\
\texttt{core} & RandomForest & 15 & 0.437 & 0.620 & n\_estimators=500, max\_depth=20, min\_samples\_leaf=5, max\_features=0.5 \\
\texttt{core} & XGBoost & 15 & 0.410 & 0.622 & n\_estimators=800, max\_depth=8, learning\_rate=0.05, min\_child\_weight=3.0, subsample=1.0, colsample\_bytree=1.0, reg\_alpha=0.0, reg\_lambda=5.0 \\
\texttt{core{+}climate} & LightGBM & 15 & 0.419 & 0.624 & n\_estimators=800, num\_leaves=127, learning\_rate=0.05, min\_child\_samples=20, reg\_alpha=0.1, reg\_lambda=5.0, colsample\_bytree=0.8, subsample=1.0, subsample\_freq=1 \\
\texttt{core{+}climate} & MLP & 15 & -0.453 & 0.657 & hidden\_layers=[128, 64, 32], dropout=0.0, lr=0.003, weight\_decay=0.0001, batch\_size=2048, max\_epochs=200, patience=20 \\
\texttt{core{+}climate} & RandomForest & 15 & 0.422 & 0.623 & n\_estimators=200, max\_depth=15, min\_samples\_leaf=5, max\_features=0.5 \\
\texttt{core{+}climate} & XGBoost & 15 & 0.412 & 0.621 & n\_estimators=800, max\_depth=10, learning\_rate=0.02, min\_child\_weight=1.0, subsample=0.7, colsample\_bytree=0.7, reg\_alpha=1.0, reg\_lambda=5.0 \\
\texttt{us\_core} & LightGBM & 15 & 0.433 & 0.576 & n\_estimators=800, num\_leaves=127, learning\_rate=0.08, min\_child\_samples=20, reg\_alpha=1.0, reg\_lambda=0.1, colsample\_bytree=1.0, subsample=0.8, subsample\_freq=1 \\
\texttt{us\_core} & MLP & 15 & 0.139 & 0.603 & hidden\_layers=[512, 256, 128], dropout=0.1, lr=0.001, weight\_decay=0.0, batch\_size=4096, max\_epochs=200, patience=20 \\
\texttt{us\_core} & RandomForest & 15 & 0.519 & 0.574 & n\_estimators=500, max\_depth=25, min\_samples\_leaf=2, max\_features=0.5 \\
\texttt{us\_core} & XGBoost & 15 & 0.451 & 0.570 & n\_estimators=800, max\_depth=10, learning\_rate=0.02, min\_child\_weight=10.0, subsample=0.7, colsample\_bytree=0.9, reg\_alpha=0.0, reg\_lambda=1.0 \\
\texttt{us\_leaky\_eui} & LightGBM & 15 & 0.763 & 0.150 & n\_estimators=800, num\_leaves=63, learning\_rate=0.1, min\_child\_samples=5, reg\_alpha=0.0, reg\_lambda=0.0, colsample\_bytree=0.7, subsample=1.0, subsample\_freq=1 \\
\texttt{us\_leaky\_eui} & MLP & 15 & 0.457 & 0.213 & hidden\_layers=[512, 512, 256], dropout=0.1, lr=0.001, weight\_decay=0.0001, batch\_size=4096, max\_epochs=200, patience=20 \\
\texttt{us\_leaky\_eui} & RandomForest & 15 & 0.789 & 0.139 & n\_estimators=200, max\_depth=25, min\_samples\_leaf=2, max\_features=0.5 \\
\texttt{us\_leaky\_eui} & XGBoost & 15 & 0.740 & 0.142 & n\_estimators=800, max\_depth=10, learning\_rate=0.1, min\_child\_weight=3.0, subsample=1.0, colsample\_bytree=1.0, reg\_alpha=1.0, reg\_lambda=1.0 \\
\texttt{us\_leaky\_full} & LightGBM & 15 & 0.794 & 0.097 & n\_estimators=800, num\_leaves=127, learning\_rate=0.1, min\_child\_samples=10, reg\_alpha=0.0, reg\_lambda=0.0, colsample\_bytree=1.0, subsample=1.0, subsample\_freq=1 \\
\texttt{us\_leaky\_full} & MLP & 15 & 0.210 & 0.117 & hidden\_layers=[512, 256, 128], dropout=0.0, lr=0.0005, weight\_decay=0.0, batch\_size=1024, max\_epochs=200, patience=20 \\
\texttt{us\_leaky\_full} & RandomForest & 15 & 0.854 & 0.076 & n\_estimators=500, max\_depth=25, min\_samples\_leaf=1, max\_features=0.7 \\
\texttt{us\_leaky\_full} & XGBoost & 15 & 0.825 & 0.085 & n\_estimators=800, max\_depth=10, learning\_rate=0.05, min\_child\_weight=10.0, subsample=0.9, colsample\_bytree=0.8, reg\_alpha=0.0, reg\_lambda=0.1 \\
\texttt{us\_metadata} & LightGBM & 15 & 0.493 & 0.470 & n\_estimators=800, num\_leaves=31, learning\_rate=0.1, min\_child\_samples=10, reg\_alpha=0.0, reg\_lambda=5.0, colsample\_bytree=0.9, subsample=0.7, subsample\_freq=1 \\
\texttt{us\_metadata} & MLP & 15 & 0.472 & 0.485 & hidden\_layers=[256, 128, 64], dropout=0.1, lr=0.001, weight\_decay=1e-05, batch\_size=2048, max\_epochs=200, patience=20 \\
\texttt{us\_metadata} & RandomForest & 15 & 0.577 & 0.453 & n\_estimators=300, max\_depth=25, min\_samples\_leaf=2, max\_features=0.5 \\
\texttt{us\_metadata} & XGBoost & 15 & 0.464 & 0.463 & n\_estimators=800, max\_depth=10, learning\_rate=0.08, min\_child\_weight=5.0, subsample=0.9, colsample\_bytree=0.8, reg\_alpha=1.0, reg\_lambda=0.1 \\
\end{longtable}
}

\section{Building-Track Feature Attribution}
\label{app:t2_feature_attribution}

Table~\ref{tab:app_t2_feature_attribution} compares permutation importance and
SHAP attribution for the tuned LightGBM model on the 26-city
\texttt{core\_all\_cities} tier. Both diagnostics identify gross floor area as
the dominant signal, followed by location; year and climate-derived covariates
are smaller on this tuned tree model, matching the main-text finding that
explicit climate fields change tuned-tree $R^2$ only marginally on the core
tier.

\begin{table}[t]
\centering
\caption{T2 Task~A feature attribution on \texttt{core\_all\_cities}, tuned LightGBM. Permutation: $R^2$ drop after shuffling (5 repeats). SHAP: mean $|\text{SHAP}|$ on 5{,}000 test rows.}
\label{tab:app_t2_feature_attribution}
\scriptsize
\setlength{\tabcolsep}{5pt}
\begin{tabular}{lcc}
\toprule
Feature & Permutation $\Delta R^2$ & Mean $|\mathrm{SHAP}|$ \\
\midrule
Gross floor area & $0.678 \pm 0.014$ & $0.729$ \\
Longitude & $0.167 \pm 0.019$ & $0.346$ \\
Latitude & $0.154 \pm 0.013$ & $0.177$ \\
Heating degree days & $0.042 \pm 0.003$ & $0.045$ \\
Reporting year & $0.033 \pm 0.005$ & $0.095$ \\
Cooling degree days & $0.025 \pm 0.003$ & $0.033$ \\
\bottomrule
\end{tabular}
\end{table}

\section{Company-Track Feature Attribution}
\label{app:t1_feature_attribution}

\begin{figure}[h]
\centering
\includegraphics[width=0.55\textwidth]{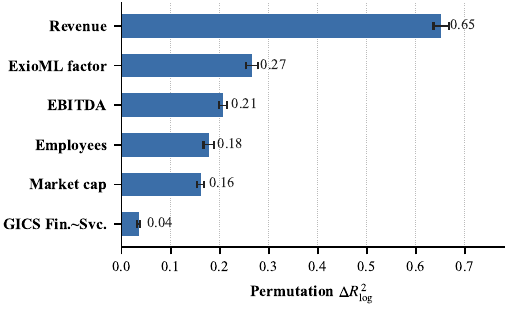}
\caption{Tuned LightGBM permutation $\Delta R^2$ on the strict-coverage panel
(top six features; error bars: std over five repeats).}
\label{fig:t1_attribution}
\end{figure}

Figure~\ref{fig:t1_attribution} and Table~\ref{tab:app_t1_feature_attribution}
report the attribution diagnostic for the tuned LightGBM model on the company
track (the strict-coverage panel). Revenue dominates with permutation
$\Delta R^2=0.652$, more than twice the next feature; the ExioML
sectoral factor is second at $0.266$, and EBITDA, employees, and market cap
form a similar-magnitude secondary group. The only sector dummy that enters
the top six is GICS Financial Services, consistent with the within-GICS
sub-sector heterogeneity discussed in the sector-factor case study
(Appendix~\ref{app:t1_sector_case}). Mean $|$SHAP$|$ rankings agree with
permutation order.

\begin{table}[t]
\centering
\caption{T1-A Scope~1+2 feature attribution on T1-Strict, tuned LightGBM. Permutation: $R^2_{\log}$ drop after shuffling (5 repeats). SHAP: mean $|\text{SHAP}|$ on full test panel. Top 6 by permutation rank.}
\label{tab:app_t1_feature_attribution}
\scriptsize
\setlength{\tabcolsep}{5pt}
\begin{tabular}{lcc}
\toprule
Feature & Permutation $\Delta R^2_{\log}$ & Mean $|\mathrm{SHAP}|$ \\
\midrule
Revenue            & $0.652 \pm 0.016$ & $0.477$ \\
ExioML factor      & $0.266 \pm 0.013$ & $0.298$ \\
EBITDA             & $0.207 \pm 0.008$ & $0.203$ \\
Employees          & $0.178 \pm 0.011$ & $0.124$ \\
Market cap         & $0.161 \pm 0.007$ & $0.110$ \\
GICS Financial Services & $0.035 \pm 0.003$ & $0.070$ \\
\bottomrule
\end{tabular}
\end{table}

\section{Compute and Runtime}
\label{app:compute}

All experiments were run on a single workstation with 8\,$\times$\,NVIDIA
RTX A5000 (24~GB) GPUs and a multi-core CPU; only TabPFN, MLP, and time-series
foundation-model inference made use of GPUs. Tree baselines (RandomForest,
XGBoost, LightGBM, HistGradientBoosting) ran exclusively on CPU.

The dominant cost is the multi-seed re-fit campaigns rather than
hyperparameter search. Indicative wall-clock numbers from the experiment logs:
\begin{itemize}[leftmargin=1.2em,itemsep=2pt,topsep=2pt]
  \item building-track regression 3-seed re-fit across the nine feature tiers and four
        model families: about 4--6 hours per seed on CPU plus
        $\le 1$\,GPU-hour per seed for MLP.
  \item Building-track regression + Sentinel-2 3-seed re-fit on the S2-eligible subset:
        about 2--3 hours per seed because the S2-aligned subset is smaller.
  \item Building-track cross-city transfer 3-seed re-fit (23 held-out cities, four model
        families): about 12--18 hours per seed on CPU.
  \item Company-track tuned tree family (40-trial search per model on the strict-coverage panel):
        $\le 1$\,CPU-hour per model.
  \item Company-track LLM panel (200-row stratified sample, four LLMs $\times$ three
        prompt regimes): a few API-hours total at standard rate limits.
\end{itemize}
A full reproduction of every multi-seed table reported in the main text
takes on the order of 80--120 CPU-core-hours plus a few GPU-hours, well
within a single-workstation budget. Per-experiment elapsed times are stored
in \texttt{elapsed\_sec} columns of the per-run CSVs and in the script logs.

\section{MLP Initialisation-only Multi-seed Refit}
\label{app:mlp_5seed}

Table~\ref{tab:app_mlp_5seed} reports a fixed-split init-only refit
(train/validation/test fold held constant; five seeds vary only the network
initialisation). This complements the headline 5-seed protocol used in
\S\ref{sec:exp_t2_feature} and Figure~\ref{fig:t2_tier_ladder_main}, where each
seed draws a fresh building-grouped split. Single-seed MLP/FT-Transformer
$R^2$ on small panels reflects optimisation variance; the per-seed table is
the more informative artefact for downstream users.

\begin{table}[t]
\centering
\caption{MLP init-only refit on T2 Task~A grouped split: 5 seeds vary network init only (split fixed). On \texttt{core\_all\_cities}, 4/5 seeds yield negative $R^2$.}
\label{tab:app_mlp_5seed}
\footnotesize
\begin{tabular}{l r r r r r r}
\toprule
Feature tier & $n_\text{seeds}$ & $R^2$ mean & $R^2$ std & $R^2$ min & $R^2$ max & MAE mean \\
\midrule
\texttt{core\_all\_cities} & 5 & -0.294 & 0.332 & -0.502 & 0.290 & 415.1 \\
\texttt{core\_all\_cities\_climate\_plus} & 5 & -0.255 & 0.337 & -0.500 & 0.340 & 408.3 \\
\texttt{us\_core} & 5 & 0.266 & 0.208 & 0.033 & 0.498 & 361.1 \\
\texttt{us\_metadata} & 5 & 0.415 & 0.154 & 0.187 & 0.543 & 302.7 \\
\texttt{us\_leaky\_eui} & 5 & 0.552 & 0.175 & 0.379 & 0.765 & 184.9 \\
\texttt{us\_leaky\_full} & 5 & 0.228 & 0.231 & -0.051 & 0.478 & 151.3 \\
\bottomrule
\end{tabular}
\end{table}

\section{Company-Track Supplement}
\label{app:t1_scope3_forecasting}

Table~\ref{tab:t1_headline} reports the full Scope~1+2 structured baseline
panel summarised by Table~\ref{tab:t1_main}. Table~\ref{tab:t1_llm_panel}
reports the LLM-only panel summarised in Section~\ref{sec:exp_t1_structured}.
Table~\ref{tab:t1_scope3_forecasting} reports the Scope~3 regression panel and
the one-step annual forecasting panel summarised in
Section~\ref{sec:exp_t1_difficulty}; the company panel spans only six
reporting years (2018--2023), which is too short to support a meaningful
multi-horizon protocol after train/val/test allocation, so company-track forecasting
is reported at a single horizon only. Multi-horizon evaluation is provided
on the longer building panel (Table~\ref{tab:app_task_e_horizon}). HistGradientBoosting hyperparameters on
Scope~3 (the matched-panel Scope~3 task) are tuned independently with a 40-trial random search on the
filtered Scope~3 panel rather than reusing the Scope~1+2 configuration; the
retuned model reaches $R^2=0.653$ in Table~\ref{tab:t1_main}.

\begin{table}[t]
\centering
\caption{T1 Scope~1+2 headline structured baselines on T1-Strict. LLM panel: Table~\ref{tab:t1_llm_panel}.}
\label{tab:t1_headline}
\footnotesize
\begin{tabular}{l l r r r r}
\toprule
Subset & Model & $n_\text{test}$ & MAE$_{\log}$ & $R^2_{\log}$ & Pearson $r$ \\
\midrule
T1-Wide & Sector-factor lookup & 1329 & 0.878 & 0.222 & 0.683 \\
 & LightGBM (default) & 3135 & 0.933 & 0.300 & 0.548 \\
 & XGBoost (default) & 3135 & 0.933 & 0.303 & 0.551 \\
\midrule
T1-Strict & Sector-factor lookup & 1329 & 0.878 & 0.222 & 0.683 \\
 & LightGBM (default) & 1363 & 0.340 & 0.826 & 0.911 \\
 & XGBoost (default) & 1363 & 0.459 & 0.732 & 0.859 \\
 & TabPFN v2 & 1363 & 0.262 & 0.857 & 0.927 \\
 & MLP (neural) & 1363 & 0.498 & 0.693 & 0.834 \\
 & FT-Transformer (neural) & 1363 & 0.565 & 0.622 & 0.790 \\
 & LightGBM (tuned) (5-seed $\sigma_{R^2}$=0.0013) & 1363 & 0.219 & 0.873 & 0.935 \\
 & XGBoost (tuned) & 1363 & 0.207 & 0.879 & 0.938 \\
 & RandomForest (tuned) & 1363 & 0.226 & 0.875 & 0.936 \\
 & HistGradientBoosting (tuned) & 1363 & 0.243 & 0.871 & 0.933 \\
\bottomrule
\end{tabular}
\end{table}

\begin{table}[t]
\centering
\caption{T1 Scope~1+2 LLM panel. Best prompt regime per model by log-MAE; paired $\Delta R^2$ = RandomForest tuned $-$ LLM on overlapping subset.}
\label{tab:t1_llm_panel}
\footnotesize
\begin{tabular}{l l r r r r r}
\toprule
LLM & Prompt & $n$ & MAE$_{\log}$ & $R^2_{\log}$ & $\Delta R^2$ vs RF & $p$ \\
\midrule
Claude Haiku 4.5 & zero-shot & 200 & 0.616 & 0.586 & 0.323 & $<0.001$ \\
GPT-4o mini & zero-shot & 200 & 0.738 & 0.491 & 0.422 & $<0.001$ \\
Qwen2.5-7B Instruct & few-shot & 200 & 0.930 & 0.213 & 0.702 & $<0.001$ \\
Mistral-7B Instruct & few-shot & 200 & 1.159 & -0.201 & 1.121 & $<0.001$ \\
\bottomrule
\end{tabular}
\end{table}

\begin{table}[t]
\centering
\caption{T1 Scope~3 prediction (tuned trees on T1-A$'$) and T1-E one-step Scope~1+2 forecasting (balanced 2018--2022 panel).}
\label{tab:t1_scope3_forecasting}
\footnotesize
\begin{tabular}{l l r r r r}
\toprule
Block & Baseline & $n_\text{test}$ & MAE$_{\log}$ & $R^2_{\log}$ & Pearson $r$ \\
\midrule
T1-A$'$ Scope~3 & LightGBM (tuned, log target) & 982 & 0.607 & 0.637 & 0.805 \\
 & XGBoost (tuned, log target) & 982 & 0.583 & 0.645 & 0.806 \\
 & RandomForest (tuned, log target) & 982 & 0.594 & 0.639 & 0.803 \\
\midrule
T1-E forecasting & Persistence & 1930 & 0.088 & 0.959 & 0.979 \\
 & Sector growth & 1930 & 0.089 & 0.959 & 0.979 \\
 & Three-year moving average & 1930 & 0.119 & 0.947 & 0.973 \\
 & Linear trend & 1930 & 0.131 & 0.936 & 0.969 \\
 & XGBoost (lag features) & 386 & 0.123 & 0.933 & 0.966 \\
 & Chronos-Bolt small & 1930 & 0.147 & 0.776 & 0.906 \\
 & TimesFM 2.0 500M & 1930 & 0.143 & 0.808 & 0.916 \\
 & Moirai 1.1 small & 1930 & 0.134 & 0.930 & 0.965 \\
\bottomrule
\end{tabular}
\end{table}

\subsection{Sector-Factor Case Study}
\label{app:t1_sector_case}

The sector-factor baseline in Table~\ref{tab:t1_main} captures coarse scale but
can miss firm-level magnitude by orders of magnitude. A representative U.K.
financial company in the panel reports \$8.5\,B revenue. The ExioML factor for
Financial Services is $10.7$\,tCO$_2$e per million USD, so the factor baseline
predicts $90{,}841$\,tCO$_2$e. The actual reported Scope~1+2 is
$750$\,tCO$_2$e, a $121\times$ over-prediction. Banks, insurers, asset managers,
and payment firms all fall under broad Financial Services labels, but their
operating-emission intensities differ substantially; a single sectoral factor
therefore cannot resolve lower-intensity sub-sectors.

Finer-grained sector labels partially reduce this error but do not close the
gap. Recomputing the intensity factor on training rows at three SICS
granularities for the same company---a private-equity asset manager---yields
$8.98$, $7.69$, and $5.15$\,tCO$_2$e per million USD at the GICS-11, SICS sector,
and SICS sub-sector (\textit{Capital Markets}) levels, reducing the
over-prediction from $121\times$ to $58\times$. Going one level finer to SICS
industry (\textit{Asset Management \& Custody Activities}, $92$ training rows)
reverses the trend to $111\times$, because the partition is too small to
estimate a stable factor. On the full $152$-row U.K. Financial Services cohort,
geometric-mean over-prediction drops from $6.9\times$ at GICS-11 to
$4.6\times$ at sub-sector and rises back to $7.5\times$ at industry. Consistent
with this pattern, adding sub-sector target encoding to the strict-coverage panel raises default
LightGBM by $\Delta R^2=+0.021$, while extending the encoding to industry
adds essentially no gain ($-0.0004$).

\section{Building-Track Transfer and Forecasting Summary}
\label{app:t2_transfer_forecasting}

Table~\ref{tab:t2_transfer_forecasting} collects the building-track stress-test
summaries: strict temporal hold-out, City-LOCO transfer, the Sentinel-2
cross-city delta, and one-step forecasting. We keep these in the appendix
because they diagnose failure modes and regime shifts rather than define the
strongest headline in-distribution baselines.

\textbf{Temporal hold-out is comparatively stable.} Task~B1 tree models retain
overall $R^2\!\approx\!0.40$--$0.50$ on \texttt{core\_all\_cities}, broadly
comparable to their grouped-building Task~A scores; strict year hold-out is
much easier than the cross-city transfer of
Section~\ref{sec:exp_t2_cross_city}.

\textbf{Forecasting has several competitive families.} On the
forecasting panel, TimesFM reaches the highest displayed $R^2$ ($0.842$),
Chronos has the lowest MAE among the displayed models ($122.1$),
RandomForest has the lowest log-MAE ($0.228$), and tree-family lag models
range from $0.705$ to $0.754$ in $R^2$. The benchmark thus surfaces three
distinct axes: cross-city transfer is difficult, strict year hold-out is
comparatively stable, and forecasting is contested across baseline
families rather than dominated by a single class.

\begin{table*}[t]
\centering
\caption{Building-track transfer and forecasting on \texttt{core\_all\_cities}. Left: temporal hold-out ($R^2$) and City-LOCO (mean$\pm$std across held-out metros$\times$seeds); +S2 is the PCA-64 Sentinel-2+Clay delta on the matched subset. Right: one-step forecasting with lagged-only inputs.}
\label{tab:t2_transfer_forecasting}
\scriptsize
\begin{minipage}[t]{0.48\linewidth}
\centering
\textbf{Temporal and cross-city}
\vspace{0.35em}

\setlength{\tabcolsep}{2.5pt}
\begin{tabular}{lccc}
\toprule
Model & Temporal & City-LOCO mean & + S2 ($\Delta$) \\
\midrule
GlobalMean & $-0.021$ & $-0.224 \pm 0.20$ & --- \\
CityTypeMean & $0.034$ & $-0.224 \pm 0.20$ & --- \\
TabPFN v2    & $0.466 \pm 0.014$ & $-0.127 \pm 1.07$ & --- \\
RF & $\mathbf{0.498}$ & $\mathbf{0.127 \pm 0.38}$ & --- \\
LightGBM     & $0.398$ & $0.030 \pm 0.66$ & $\mathbf{+0.124}$ \\
XGBoost      & $0.389$ & $0.084 \pm 0.60$ & $+0.053$ \\
MLP          & $0.368$ & $-2.55 \pm 9.83$ & $-0.229$ \\
\bottomrule
\end{tabular}
\end{minipage}
\hfill
\begin{minipage}[t]{0.48\linewidth}
\centering
\textbf{One-step forecasting}
\vspace{0.35em}

\setlength{\tabcolsep}{2.5pt}
\begin{tabular}{lccc}
\toprule
Model & MAE & LogMAE & $R^2$ \\
\midrule
GlobalMean & $543.7$ & $1.164$ & $-0.021$ \\
CityTypeMean & $495.1$ & $0.894$ & $0.048$ \\
LinearTrend & $172.4$ & $0.397$ & $0.616$ \\
RF & $131.7$ & $\mathbf{0.228}$ & $0.754$ \\
LightGBM & $143.7$ & $0.233$ & $0.715$ \\
XGBoost & $144.9$ & $0.231$ & $0.705$ \\
MLP & $322.4$ & $0.822$ & $0.555$ \\
Chronos & $\mathbf{122.1}$ & $0.235$ & $0.819$ \\
TimesFM & $125.5$ & $0.236$ & $\mathbf{0.842}$ \\
Moirai & $139.1$ & $0.234$ & $0.783$ \\
\bottomrule
\end{tabular}
\end{minipage}
\end{table*}

\textbf{Multi-horizon forecasting.} The Task~E protocol fixes train end
at year~2019 but evaluates every test year $y\!\geq\!2021$ separately,
so the same trained model is queried at horizons $h\!=\!y\!-\!2019$
ranging from $h\!=\!2$ (2021) to $h\!=\!5$ (2024). Per-year $R^2$ is
in Table~\ref{tab:app_task_e_horizon}. Time-series foundation models
(TimesFM, Chronos) hold up best at the longer horizons, reaching
$R^2\!=\!0.90$ at $h\!=\!4$; lag-feature gradient-boosted regressors
plateau slightly lower ($0.66$--$0.81$); MLP degrades fastest
($0.54$ at $h\!=\!2$ to $0.39$ at $h\!=\!5$). The horizon decay is
mild because the lag-feature input remains the immediately preceding
year (i.e.\ each prediction is one-step from its own past), so what is
varied is mainly distributional drift between train end and test year
rather than recursive multi-step error accumulation.

\begin{table}[t]
\centering
\caption{T2 Task~E multi-horizon forecasting on \texttt{core\_all\_cities}: train on years $\le$2019, validate on 2020, evaluate each test year $y$ at horizon $h=y-2019$. Each cell is per-year $R^2$. Trees and TabPFN-style baselines use lagged emissions and lagged climate; foundation models (Chronos / TimesFM / Moirai) consume the lagged emissions sequence directly. Ridge collapses on this lag-feature setup at every horizon (heavy negative; omitted from comparison).}
\label{tab:app_task_e_horizon}
\footnotesize
\setlength{\tabcolsep}{4pt}
\begin{tabular}{l c c c c}
\toprule
Model & $h\!=\!2$ (2021) & $h\!=\!3$ (2022) & $h\!=\!4$ (2023) & $h\!=\!5$ (2024) \\
\midrule
\multicolumn{5}{l}{\textit{Naive references}} \\
GlobalMean   & $-0.02$ & $-0.02$ & $-0.03$ & $-0.02$ \\
CityTypeMean & $+0.06$ & $+0.05$ & $+0.05$ & $+0.02$ \\
LinearTrend  & $+0.61$ & $+0.67$ & $+0.47$ & $+0.73$ \\
\midrule
\multicolumn{5}{l}{\textit{Time-series foundation models}} \\
Chronos      & $+0.73$ & $+0.84$ & $\mathbf{+0.90}$ & $+0.81$ \\
TimesFM      & $\mathbf{+0.77}$ & $\mathbf{+0.88}$ & $\mathbf{+0.90}$ & $+0.82$ \\
Moirai       & $+0.74$ & $+0.72$ & $+0.86$ & $+0.82$ \\
\midrule
\multicolumn{5}{l}{\textit{Lag-feature regressors}} \\
RandomForest & $+0.68$ & $+0.78$ & $+0.81$ & $+0.74$ \\
LightGBM     & $+0.67$ & $+0.74$ & $+0.77$ & $+0.67$ \\
XGBoost      & $+0.66$ & $+0.72$ & $+0.78$ & $+0.64$ \\
MLP          & $+0.54$ & $+0.66$ & $+0.59$ & $+0.39$ \\
\bottomrule
\end{tabular}
\end{table}

\section{Task B Full Model Grid}
\label{app:task_b_grid}

The main text reports Task B as a brief observation: strict year hold-out
does not collapse on \texttt{core\_all\_cities}, with tree models retaining
overall $R^2$ in the $0.40$--$0.50$ range. Table~\ref{tab:app_task_b}
provides the complete model grid for both the B1 strict-year hold-out and
the B2 COVID variant, including the auxiliary baselines that the main text
omits. The B2 COVID variant trains on years $\le 2018$ and evaluates 2019,
2020, and 2021 separately, isolating pandemic-year disruptions from
in-distribution temporal drift.

\begin{table}[t]
\centering
\caption{Task B on \texttt{core\_all\_cities}. B1: train $\le 2019$, val 2020, test $\ge 2021$. B2: train $\le 2018$, evaluate 2019/2020/2021 separately.}
\label{tab:app_task_b}
\footnotesize
\begin{tabular}{l r r r r r r}
\toprule
Model & B1 $n_\text{train}$ & B1 $n_\text{test}$ & B1 $R^2$ & B2 $R^2_{2019}$ & B2 $R^2_{2020}$ & B2 $R^2_{2021}$ \\
\midrule
CityTypeMean & 212003 & 212279 & 0.034 & 0.061 & 0.072 & 0.050 \\
GlobalMean & 212003 & 212279 & -0.021 & -0.023 & -0.020 & -0.015 \\
LightGBM & 212003 & 212279 & 0.398 & 0.512 & 0.488 & 0.416 \\
MLP & 212003 & 212279 & 0.368 & 0.426 & 0.371 & 0.256 \\
RandomForest & 212003 & 212279 & 0.498 & 0.596 & 0.568 & 0.495 \\
Ridge & 212003 & 212279 & -1.322 & -1.521 & -2.301 & -1.582 \\
XGBoost & 212003 & 212279 & 0.389 & 0.502 & 0.472 & 0.400 \\
\bottomrule
\end{tabular}
\end{table}

\begin{figure}[t]
  \centering
  \includegraphics[width=\linewidth]{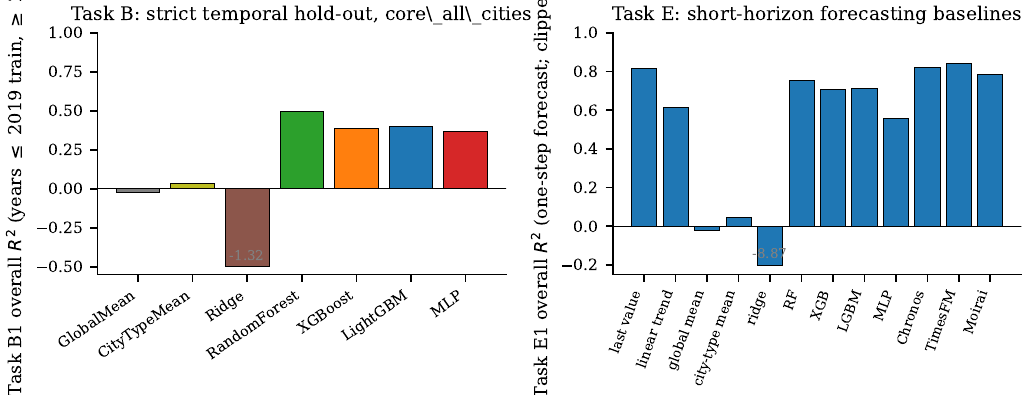}
  \caption{Left: Task~B1 strict temporal hold-out $R^2$ on
  \texttt{core\_all\_cities} (single run). Right: Task~E1 short-horizon
  forecasting $R^2$. Both panels clipped on the negative side; raw Ridge
  values annotated.}
  \label{fig:t2_temporal}
\end{figure}

\section{Task D Cross-Property-Type Stress Test}
\label{app:task_d}

Per Section~\ref{sec:exp_t2_feature}, Task D is reported as appendix-level
evidence. Table~\ref{tab:app_task_d} lists the leave-one-property-type-out
performance on the seven main types (Office, Multifamily Housing, Retail,
Hotel, K-12 School, Hospital/Medical, Warehouse/Distribution) for the
\texttt{us\_core} and \texttt{us\_metadata} tiers, which are the two tiers
with reliable property-type coverage across U.S. cities. Performance varies
substantially by held-out type, and the metadata tier does not uniformly
help every type.

{\footnotesize
\begin{longtable}{l l l r r r}
\caption{Task D cross-property-type stress test on \texttt{us\_core} and \texttt{us\_metadata}: hold out one type at training, evaluate on it.}
\label{tab:app_task_d} \\
\toprule
Tier & Held-out type & Model & $n_\text{test}$ & $R^2$ & LogMAE \\
\midrule
\endfirsthead
\multicolumn{6}{c}{\tablename\ \thetable\ -- continued} \\
\toprule
Tier & Held-out type & Model & $n_\text{test}$ & $R^2$ & LogMAE \\
\midrule
\endhead
\bottomrule
\endfoot
\texttt{us\_core} & Office & CityTypeMean & 45650 & -1.202 & 1.606 \\
\texttt{us\_core} & Office & GlobalMean & 45650 & -0.148 & 1.560 \\
\texttt{us\_core} & Office & LightGBM & 45650 & 0.245 & 0.939 \\
\texttt{us\_core} & Office & MLP & 45650 & 0.033 & 1.119 \\
\texttt{us\_core} & Office & RandomForest & 45650 & 0.392 & 1.017 \\
\texttt{us\_core} & Office & Ridge & 45650 & -1.132 & 1.409 \\
\texttt{us\_core} & Office & TabPFN v2 & 45650 & 0.298 & 0.851 \\
\texttt{us\_core} & Office & XGBoost & 45650 & 0.254 & 0.943 \\
\texttt{us\_core} & Multifamily Housing & CityTypeMean & 256168 & -1.755 & 0.996 \\
\texttt{us\_core} & Multifamily Housing & GlobalMean & 256168 & -0.048 & 1.011 \\
\texttt{us\_core} & Multifamily Housing & LightGBM & 256168 & 0.230 & 0.799 \\
\texttt{us\_core} & Multifamily Housing & MLP & 256168 & 0.069 & 1.059 \\
\texttt{us\_core} & Multifamily Housing & RandomForest & 256168 & 0.340 & 0.868 \\
\texttt{us\_core} & Multifamily Housing & Ridge & 256168 & -6.400 & 0.953 \\
\texttt{us\_core} & Multifamily Housing & TabPFN v2 & 256168 & 0.296 & 0.739 \\
\texttt{us\_core} & Multifamily Housing & XGBoost & 256168 & 0.218 & 0.794 \\
\texttt{us\_core} & Retail & CityTypeMean & 8152 & -0.915 & 1.358 \\
\texttt{us\_core} & Retail & GlobalMean & 8152 & -0.068 & 1.402 \\
\texttt{us\_core} & Retail & LightGBM & 8152 & 0.269 & 1.084 \\
\texttt{us\_core} & Retail & MLP & 8152 & 0.119 & 1.111 \\
\texttt{us\_core} & Retail & RandomForest & 8152 & 0.316 & 1.169 \\
\texttt{us\_core} & Retail & Ridge & 8152 & 0.107 & 1.314 \\
\texttt{us\_core} & Retail & TabPFN v2 & 8152 & 0.350 & 1.031 \\
\texttt{us\_core} & Retail & XGBoost & 8152 & 0.186 & 1.098 \\
\texttt{us\_core} & Hotel & CityTypeMean & 9959 & -1.636 & 1.528 \\
\texttt{us\_core} & Hotel & GlobalMean & 9959 & -0.136 & 1.300 \\
\texttt{us\_core} & Hotel & LightGBM & 9959 & 0.183 & 1.010 \\
\texttt{us\_core} & Hotel & MLP & 9959 & 0.047 & 1.180 \\
\texttt{us\_core} & Hotel & RandomForest & 9959 & 0.224 & 1.144 \\
\texttt{us\_core} & Hotel & Ridge & 9959 & -0.503 & 1.364 \\
\texttt{us\_core} & Hotel & TabPFN v2 & 9959 & 0.309 & 0.624 \\
\texttt{us\_core} & Hotel & XGBoost & 9959 & 0.164 & 1.016 \\
\texttt{us\_core} & K-12 School & CityTypeMean & 24657 & -7.703 & 1.126 \\
\texttt{us\_core} & K-12 School & GlobalMean & 24657 & -0.091 & 0.971 \\
\texttt{us\_core} & K-12 School & LightGBM & 24657 & 0.096 & 0.696 \\
\texttt{us\_core} & K-12 School & MLP & 24657 & 0.040 & 0.787 \\
\texttt{us\_core} & K-12 School & RandomForest & 24657 & 0.146 & 0.746 \\
\texttt{us\_core} & K-12 School & Ridge & 24657 & -0.075 & 1.051 \\
\texttt{us\_core} & K-12 School & TabPFN v2 & 24657 & -0.011 & 0.773 \\
\texttt{us\_core} & K-12 School & XGBoost & 24657 & 0.092 & 0.689 \\
\texttt{us\_core} & Hospital/Medical & CityTypeMean & 5282 & -0.168 & 2.098 \\
\texttt{us\_core} & Hospital/Medical & GlobalMean & 5282 & -0.166 & 2.021 \\
\texttt{us\_core} & Hospital/Medical & LightGBM & 5282 & 0.002 & 1.685 \\
\texttt{us\_core} & Hospital/Medical & MLP & 5282 & -0.073 & 1.869 \\
\texttt{us\_core} & Hospital/Medical & RandomForest & 5282 & 0.189 & 1.663 \\
\texttt{us\_core} & Hospital/Medical & Ridge & 5282 & -0.219 & 1.915 \\
\texttt{us\_core} & Hospital/Medical & TabPFN v2 & 5282 & 0.153 & 1.227 \\
\texttt{us\_core} & Hospital/Medical & XGBoost & 5282 & 0.002 & 1.681 \\
\texttt{us\_core} & Warehouse/Distribution & CityTypeMean & 13497 & -13.410 & 1.639 \\
\texttt{us\_core} & Warehouse/Distribution & GlobalMean & 13497 & -0.003 & 1.705 \\
\texttt{us\_core} & Warehouse/Distribution & LightGBM & 13497 & 0.169 & 0.973 \\
\texttt{us\_core} & Warehouse/Distribution & MLP & 13497 & 0.154 & 1.013 \\
\texttt{us\_core} & Warehouse/Distribution & RandomForest & 13497 & 0.340 & 1.030 \\
\texttt{us\_core} & Warehouse/Distribution & Ridge & 13497 & 0.135 & 1.150 \\
\texttt{us\_core} & Warehouse/Distribution & TabPFN v2 & 13497 & -0.673 & 1.767 \\
\texttt{us\_core} & Warehouse/Distribution & XGBoost & 13497 & 0.182 & 0.975 \\
\texttt{us\_metadata} & Office & CityTypeMean & 45650 & -1.202 & 1.606 \\
\texttt{us\_metadata} & Office & GlobalMean & 45650 & -0.148 & 1.560 \\
\texttt{us\_metadata} & Office & LightGBM & 45650 & 0.291 & 1.009 \\
\texttt{us\_metadata} & Office & MLP & 45650 & 0.036 & 1.183 \\
\texttt{us\_metadata} & Office & RandomForest & 45650 & 0.386 & 1.092 \\
\texttt{us\_metadata} & Office & Ridge & 45650 & -0.983 & 1.498 \\
\texttt{us\_metadata} & Office & TabPFN v2 & 45650 & 0.364 & 0.734 \\
\texttt{us\_metadata} & Office & XGBoost & 45650 & 0.319 & 1.010 \\
\texttt{us\_metadata} & Multifamily Housing & CityTypeMean & 256168 & -1.755 & 0.996 \\
\texttt{us\_metadata} & Multifamily Housing & GlobalMean & 256168 & -0.048 & 1.011 \\
\texttt{us\_metadata} & Multifamily Housing & LightGBM & 256168 & 0.398 & 0.885 \\
\texttt{us\_metadata} & Multifamily Housing & MLP & 256168 & 0.161 & 1.158 \\
\texttt{us\_metadata} & Multifamily Housing & RandomForest & 256168 & 0.447 & 0.962 \\
\texttt{us\_metadata} & Multifamily Housing & Ridge & 256168 & -7.240 & 0.910 \\
\texttt{us\_metadata} & Multifamily Housing & TabPFN v2 & 256168 & 0.415 & 0.583 \\
\texttt{us\_metadata} & Multifamily Housing & XGBoost & 256168 & 0.401 & 0.884 \\
\texttt{us\_metadata} & Retail & CityTypeMean & 8152 & -0.915 & 1.358 \\
\texttt{us\_metadata} & Retail & GlobalMean & 8152 & -0.068 & 1.402 \\
\texttt{us\_metadata} & Retail & LightGBM & 8152 & 0.254 & 1.247 \\
\texttt{us\_metadata} & Retail & MLP & 8152 & 0.181 & 1.164 \\
\texttt{us\_metadata} & Retail & RandomForest & 8152 & 0.274 & 1.327 \\
\texttt{us\_metadata} & Retail & Ridge & 8152 & 0.107 & 1.293 \\
\texttt{us\_metadata} & Retail & TabPFN v2 & 8152 & 0.373 & 0.993 \\
\texttt{us\_metadata} & Retail & XGBoost & 8152 & 0.260 & 1.236 \\
\texttt{us\_metadata} & Hotel & CityTypeMean & 9959 & -1.636 & 1.528 \\
\texttt{us\_metadata} & Hotel & GlobalMean & 9959 & -0.136 & 1.300 \\
\texttt{us\_metadata} & Hotel & LightGBM & 9959 & 0.320 & 0.987 \\
\texttt{us\_metadata} & Hotel & MLP & 9959 & 0.129 & 1.206 \\
\texttt{us\_metadata} & Hotel & RandomForest & 9959 & 0.303 & 1.028 \\
\texttt{us\_metadata} & Hotel & Ridge & 9959 & -0.463 & 1.288 \\
\texttt{us\_metadata} & Hotel & TabPFN v2 & 9959 & 0.348 & 0.605 \\
\texttt{us\_metadata} & Hotel & XGBoost & 9959 & 0.307 & 0.984 \\
\texttt{us\_metadata} & K-12 School & CityTypeMean & 24657 & -7.703 & 1.126 \\
\texttt{us\_metadata} & K-12 School & GlobalMean & 24657 & -0.091 & 0.971 \\
\texttt{us\_metadata} & K-12 School & LightGBM & 24657 & 0.209 & 0.703 \\
\texttt{us\_metadata} & K-12 School & MLP & 24657 & 0.144 & 0.735 \\
\texttt{us\_metadata} & K-12 School & RandomForest & 24657 & 0.218 & 0.727 \\
\texttt{us\_metadata} & K-12 School & Ridge & 24657 & -0.571 & 1.058 \\
\texttt{us\_metadata} & K-12 School & TabPFN v2 & 24657 & -0.114 & 0.894 \\
\texttt{us\_metadata} & K-12 School & XGBoost & 24657 & 0.200 & 0.709 \\
\texttt{us\_metadata} & Hospital/Medical & CityTypeMean & 5282 & -0.168 & 2.098 \\
\texttt{us\_metadata} & Hospital/Medical & GlobalMean & 5282 & -0.166 & 2.021 \\
\texttt{us\_metadata} & Hospital/Medical & LightGBM & 5282 & 0.158 & 1.766 \\
\texttt{us\_metadata} & Hospital/Medical & MLP & 5282 & -0.040 & 1.942 \\
\texttt{us\_metadata} & Hospital/Medical & RandomForest & 5282 & 0.241 & 1.815 \\
\texttt{us\_metadata} & Hospital/Medical & Ridge & 5282 & -0.147 & 1.896 \\
\texttt{us\_metadata} & Hospital/Medical & TabPFN v2 & 5282 & 0.173 & 1.151 \\
\texttt{us\_metadata} & Hospital/Medical & XGBoost & 5282 & 0.147 & 1.768 \\
\texttt{us\_metadata} & Warehouse/Distribution & CityTypeMean & 13497 & -13.410 & 1.639 \\
\texttt{us\_metadata} & Warehouse/Distribution & GlobalMean & 13497 & -0.003 & 1.705 \\
\texttt{us\_metadata} & Warehouse/Distribution & LightGBM & 13497 & 0.461 & 0.754 \\
\texttt{us\_metadata} & Warehouse/Distribution & MLP & 13497 & 0.329 & 0.737 \\
\texttt{us\_metadata} & Warehouse/Distribution & RandomForest & 13497 & 0.357 & 0.972 \\
\texttt{us\_metadata} & Warehouse/Distribution & Ridge & 13497 & 0.067 & 1.062 \\
\texttt{us\_metadata} & Warehouse/Distribution & TabPFN v2 & 13497 & -0.803 & 1.805 \\
\texttt{us\_metadata} & Warehouse/Distribution & XGBoost & 13497 & 0.450 & 0.749 \\
\end{longtable}
}

\section{Paired-bootstrap and Per-city Tables}
\label{app:paired}

\begin{figure}[t]
  \centering
  \includegraphics[width=\linewidth]{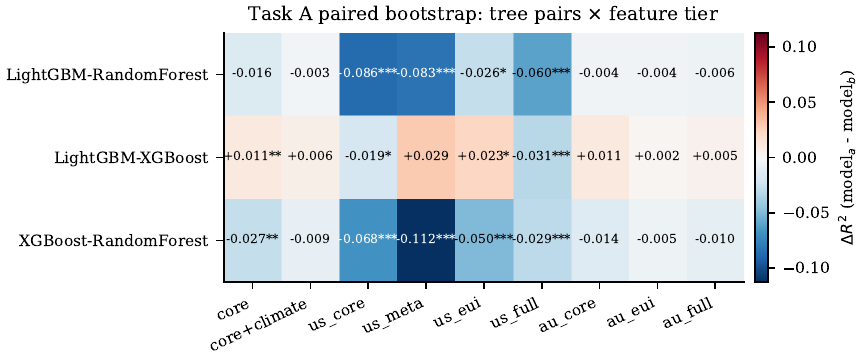}
  \caption{Paired-bootstrap $\Delta R^2$ between tree-family pairs per
  feature tier. Stars: $p_{R^2}<0.05$ ($\ast$), $<0.01$ ($\ast\ast$),
  $<0.001$ ($\ast{\ast}{\ast}$).}
  \label{fig:t2_paired_bootstrap}
\end{figure}

Full paired-bootstrap deltas and $p$-values for every (feature tier, model
pair) on building-track regression are released as the canonical CSVs
\texttt{results/clean\_building/task\_a\_pairwise\_bootstrap.csv} (US and
cross-country, tree-vs-tree),
\texttt{task\_a\_pairwise\_bootstrap\_au.csv} (AU, tree-vs-tree), and
\texttt{task\_a\_pairwise\_bootstrap\_with\_tabpfn.csv} (TabPFN~v2 vs each
tuned tree, all six tiers; Table~\ref{tab:app_t2_paired}).
Table~\ref{tab:app_t2_paired} summarises the TabPFN-vs-tree comparison
that backs the main-text claim in \S\ref{sec:exp_t1_sector_factor}.
Figure~\ref{fig:t2_paired_bootstrap} visualises the $\Delta R^2$ matrix
together with significance stars; the CSVs additionally expose paired
deltas on MAE and log-MAE, the win-rate fractions on bootstrap draws, and
the bootstrap $n_\text{test}$ used per cell.

\textbf{RandomForest's grouped-split advantage is schema-dependent, not
universal.} On the four U.S.-only tiers from clean metadata through
target-proxy fields, RandomForest exceeds LightGBM by $0.026$--$0.086$~$R^2$
with $p\!\leq\!0.018$, and exceeds XGBoost on three of those four tiers
at $p<0.05$. On the two cross-country tiers the effect is smaller and less
consistent, and on all three AU-only tiers every tree-vs-tree pair has
$p_{R^2}\!\geq\!0.088$. The natural reading is that RandomForest's bagging
advantage is strongest inside the heterogeneous U.S. metadata ladder,
weakens under deliberately minimal cross-country schemas, and disappears
on Australian disclosures, which share a single NABERS/BEEC schema.

Per-city stratified $R^2$ for building-track regression is released as
\texttt{task\_a\_stratified\_r2.csv} (US-only) and
\texttt{task\_a\_stratified\_r2\_au.csv} (AU-only); each row is a (feature
tier, model, city) triple with three-seed mean and standard deviation. The
522 rows are too long for a paper table, so they remain in the released CSVs.

Table~\ref{tab:app_task_c1_city_r2} reports the per-held-out-city $R^2$
values behind the main-text City-LOCO summary on the
\texttt{core\_all\_cities} tier (Figure~\ref{fig:t2_cross_city}). The table
keeps the raw pathological values for all models including tabular MLP,
which is omitted from the main figure.

\begin{table}[t]
\centering
\caption{Per-held-out-city $R^2$ for T2 Task~C1 City-LOCO on
\texttt{core\_all\_cities}, 3-seed mean.}
\label{tab:app_task_c1_city_r2}
\scriptsize
\setlength{\tabcolsep}{3.5pt}
\begin{tabular}{lrrrr}
\toprule
Held-out city & RF & LightGBM & XGBoost & MLP \\
\midrule
Adelaide & $0.089$ & $0.051$ & $-0.009$ & $0.153$ \\
Boston & $0.169$ & $0.009$ & $0.017$ & $0.398$ \\
Brisbane & $0.376$ & $0.485$ & $0.440$ & $0.486$ \\
Cairns & $0.518$ & $0.677$ & $0.693$ & $0.462$ \\
Canberra & $0.192$ & $0.149$ & $0.177$ & $-0.130$ \\
Chicago & $0.216$ & $0.257$ & $0.452$ & $0.407$ \\
Darwin & $0.249$ & $-0.697$ & $-0.202$ & $-0.623$ \\
DC & $0.230$ & $0.207$ & $0.198$ & $0.276$ \\
Denver & $0.275$ & $0.371$ & $0.375$ & $0.305$ \\
Gold Coast & $0.233$ & $0.262$ & $0.277$ & $0.269$ \\
Hobart & $-0.601$ & $-2.511$ & $-2.511$ & $0.258$ \\
Los Angeles & $0.120$ & $0.019$ & $-0.100$ & $-0.085$ \\
Melbourne & $0.088$ & $0.040$ & $0.052$ & $-0.088$ \\
Newcastle & $0.548$ & $0.534$ & $0.541$ & $0.483$ \\
New York City & $0.279$ & $0.210$ & $0.204$ & $0.272$ \\
Perth & $0.406$ & $0.436$ & $0.423$ & $-36.303$ \\
Philadelphia & $-0.099$ & $0.086$ & $0.076$ & $0.097$ \\
Portland & $0.340$ & $0.164$ & $0.143$ & $0.173$ \\
Seattle & $-1.227$ & $-0.972$ & $-0.131$ & $-0.910$ \\
San Francisco & $0.450$ & $0.343$ & $0.187$ & $0.186$ \\
Singapore & $-0.087$ & $-0.017$ & $-0.011$ & $-0.163$ \\
Sydney & $0.187$ & $0.193$ & $0.210$ & $-27.290$ \\
Townsville & $0.303$ & $0.322$ & $0.301$ & $0.275$ \\
Wollongong & $-0.213$ & $0.111$ & $0.213$ & $-0.023$ \\
\bottomrule
\end{tabular}
\end{table}

\begin{table}[t]
\centering
\caption{T2 Task~A paired bootstrap (1{,}000 resamples on shared
grouped-building test rows, seed~42): TabPFN~v2 minus each tuned tree.
Positive $\Delta R^2$ favours TabPFN. $p$-values are two-sided. Bold
entries are significant at $p<.05$.}
\label{tab:app_t2_paired}
\small
\setlength{\tabcolsep}{4pt}
\begin{tabular}{lrrrr}
\toprule
& & \multicolumn{3}{c}{$\Delta R^2$ (TabPFN $-$ tree) [95\% CI], $p$} \\
\cmidrule(lr){3-5}
Tier & $n_{\text{test}}$ & vs LightGBM & vs XGBoost & vs RandomForest \\
\midrule
core (26 cities)            & 94{,}875 & \textbf{+.025 [+.005, +.044], .012} & \textbf{+.036 [+.014, +.058], .002} & +.010 [$-$.011, +.028], .362 \\
core+climate (26 cities)    & 94{,}875 & \textbf{+.033 [+.018, +.046], .000} & \textbf{+.039 [+.021, +.056], .000} & \textbf{+.029 [+.015, +.043], .000} \\
us\_core (6 cities)         & 86{,}330 & \textbf{+.109 [+.075, +.143], .000} & \textbf{+.090 [+.061, +.121], .000} & \textbf{+.022 [+.001, +.044], .040} \\
us\_metadata (6 cities)     & 86{,}330 & \textbf{+.108 [+.079, +.140], .000} & \textbf{+.137 [+.088, +.195], .000} & \textbf{+.024 [+.005, +.044], .016} \\
au\_core (13 cities)        & 3{,}704  & $-$.017 [$-$.108, +.041], .826      & $-$.006 [$-$.096, +.053], .998      & $-$.020 [$-$.110, +.037], .740 \\
au\_eui (13 cities)         & 3{,}704  & +.005 [$-$.077, +.054], .762        & +.007 [$-$.075, +.058], .744        & +.001 [$-$.084, +.050], .852 \\
\bottomrule
\end{tabular}
\end{table}

\section{Sentinel-2 + Clay Supplementary Variants}
\label{app:s2_supp}

\begin{figure}[t]
  \centering
  \includegraphics[width=\linewidth]{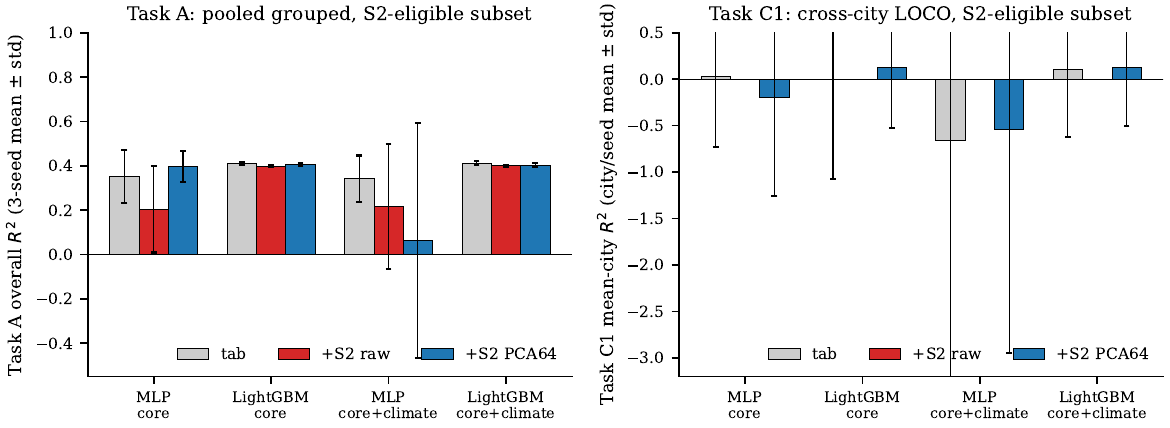}
  \caption{Sentinel-2 + Clay multimodal extension. Left: Task~A grouped
  $R^2$ (3-seed mean$\pm$std) for tab, tab+raw, and tab+PCA-64. Right:
  Task~C1 mean-city $R^2$ across held-out cities and seeds for the same
  variants.}
  \label{fig:t2_s2_gain}
\end{figure}

The headline Sentinel-2 figure (Figure~\ref{fig:t2_s2_gain}) excludes the
PCA-128 and gated two-tower fusion variants because they are either
single-seed exploratory or do not have multi-seed coverage on both Task~A
and Task~C1. Single-seed numbers for these variants are released as
\texttt{task\_a\_results\_s2\_all.csv} and
\texttt{task\_c\_results\_s2\_all.csv}; per-city S2 gain deltas under LOCO
are released as \texttt{task\_c1\_s2\_3seeds\_city\_deltas.csv} so that
readers interested in the heterogeneous geographic effect of Sentinel-2 can
inspect the per-target-city deltas directly.

\section{LLM Panel: Prompt Templates and Failure Modes}
\label{app:llm_prompts}

The Company-track LLM panel uses one shared prompt scaffold across three regimes
(\textit{zero-shot}, \textit{few-shot}, \textit{no-summary}), constructed by
\texttt{scripts/run\_t1\_a\_llm.py}. The scaffold has four blocks: a system
instruction, an optional labelled-example block (few-shot only), a target
company description, and a closing line that invites a single numeric answer.

\textbf{System instruction (verbatim).}
\begin{quote}\small\ttfamily
You are an expert in corporate greenhouse-gas accounting. Given the company
information below, predict Scope 1 + Scope 2 emissions in metric tons of
CO2-equivalent (tCO2e) for the reporting year.\\[2pt]
Respond with a SINGLE NUMBER ONLY --- the predicted tCO2e value. Do not
include units, explanations, or punctuation. Example: 42500
\end{quote}

\textbf{Company description block.} Each company line set covers \textit{Company}, \textit{Country (ISO-2)},
\textit{Sector (GICS-11)}, \textit{Reporting year}, \textit{Annual revenue}
(in millions of USD with FX correction), and an optional 800-character
\textit{Business description}. Few-shot mode prepends ten labelled examples
drawn one per GICS-11 sector from a held-out fewshot pool, each followed by
the literal line \texttt{Scope 1+2 emissions: \{value\} tCO2e}. The
\textit{no-summary} regime drops the business-description line.

\textbf{Parsing and failure modes.} The response is parsed by extracting the first numeric token (regular
expression \texttt{(-?[\textbackslash d,]+\textbackslash .?\textbackslash d*)}
with optional scientific notation), stripping commas, and casting to a float.
Predictions that fail to parse are flagged \texttt{NaN} and excluded from the
test-set metric computation. The most common failure modes observed during
runs are: (i) the LLM emits a sentence rather than a number when the
description is unusually short; (ii) the LLM returns a range
(e.g.\ \texttt{40000--50000}); (iii) the LLM returns scientific notation
without an exponent (e.g.\ \texttt{4.5e}). The parser tolerates (iii) and
recovers the numeric prefix, but (i) and (ii) yield \texttt{NaN}; the
fraction of \texttt{NaN} responses per (model, regime) is recorded in the
per-prediction CSVs released with the benchmark.

\section{Reproducibility and Release}
\label{app:reproducibility}

The project repository contains the executable reconstruction and evaluation
code, canonical split definitions, feature-tier registries, harmonisation
metadata, and scripts for regenerating the paper figures and tables from the
canonical CSVs. Source data are retrieved from their original portals under
the licenses listed in Table~\ref{tab:data_sources}. Reproducing the full
benchmark therefore involves: (i) running the data acquisition step for each
portal under its own access terms, (ii) running the canonical preprocessing and
feature-tier construction, (iii) running the experiment scripts under
\texttt{scripts/} for each track and task, and (iv) regenerating the figures
and tables with \texttt{scripts/plots/}, which read only from the canonical
CSVs and emit deterministic outputs.

\section{Source Access and License Notes}
\label{app:data_sources}

Table~\ref{tab:data_sources} records the source/access plan used for benchmark
reconstruction. Raw records are retrieved from their original portals and APIs
rather than redistributed wholesale.

\begin{table}[t]
\centering
\caption{Source access and license notes. Raw records are retrieved from the original sources under the listed terms; we release code, harmonisation metadata, and reconstruction recipes.}
\label{tab:data_sources}
\scriptsize
\resizebox{\linewidth}{!}{%
\begin{tabular}{p{0.17\linewidth} p{0.20\linewidth} p{0.12\linewidth} p{0.26\linewidth} p{0.27\linewidth}}
\toprule
Source & Coverage in GHGbench & Years used & Terms / license checked & Access method \\
\midrule
NYC LL84 benchmarking & New York City buildings & 2012--2024 & NYC Open Data / NYC.gov terms of use & NYC Open Data portal / SODA API \\
Chicago benchmarking & Chicago buildings & 2014--2023 & City data portal open-data terms; dataset metadata did not expose a more specific license in the local archive & Chicago Data Portal CSV export \\
Seattle benchmarking & Seattle buildings & 2015--2024 & City of Seattle open-data program and website terms & Seattle Open Data portal CSV export \\
Washington DC benchmarking & District of Columbia buildings & 2013--2024 & District Data terms; District logos excluded from CC0-style designation & Open Data DC CSV export \\
San Francisco benchmarking & San Francisco buildings & 2011--2025 & DataSF public-domain dedication / PDDL policy & DataSF CSV export \\
Los Angeles EBEWE & Los Angeles buildings & 2016--2025 & DataLA open-data portal terms of use & Los Angeles Open Data CSV export \\
Boston BERDO & Boston buildings & 2016--2024 & Analyze Boston public-domain dedication / PDDL default & Analyze Boston CSV/XLSX exports \\
Denver benchmarking & Denver buildings & 2025 & Denver Open Data Catalog CC BY 3.0 terms & Denver open-data CSV export \\
Philadelphia benchmarking & Philadelphia buildings & 2023 & City of Philadelphia website / open-data terms & Philadelphia open-data CSV export \\
Portland benchmarking & Portland buildings & 2019, 2023--2024 & Portland open-data terms; PDDL used for open-data publications where specified & Portland open-data XLSX exports \\
Australian Commercial Building Disclosure & 15 Australian metropolitan areas & 2011--2026 & Creative Commons Attribution 3.0 Australia on data.gov.au CBD dataset & data.gov.au / CBD downloadable dataset \\
Singapore BCA building energy performance & Singapore buildings & 2017--2020 & Singapore Open Data Licence via data.gov.sg / BCA publication & BCA / data.gov.sg CSV/XLSX exports \\
NASA POWER & Building-year meteorology and solar covariates for T2 & 2011--2025 & NASA science data terms; NASA-led data are CC0 unless marked otherwise & NASA POWER daily point API \\
\midrule
CDU & Company Scope~1+2 and Scope~3 disclosures for T1 & 2018--2023 & CDU Terms of Service; non-commercial and download/session restrictions apply & Reconstruction recipe against CDU exports/API \\
ExioML / EXIOBASE & Sectoral emission-factor lookup for T1 & 1995--2022 source panel; matched to T1 reporting years & EXIOBASE 3 licence information is version-specific on Zenodo; current EXIOBASE3 releases use non-commercial academic terms & Local ExioML factor-accounting files and lookup script \\
Sentinel-2 + Clay & T2 satellite embeddings & Sentinel-2 L2A-era image composites aligned to S2-eligible building rows & Copernicus Sentinel free/full/open access terms; Clay model Apache-2.0 & Sentinel-2 retrieval + Clay embedding extraction pipeline \\
\bottomrule
\end{tabular}%
}
\end{table}

\clearpage
\section*{NeurIPS Paper Checklist}

\begin{enumerate}

\item {\bf Claims}
    \item[] Question: Do the main claims made in the abstract and introduction accurately reflect the paper's contributions and scope?
    \item[] Answer: \answerYes{} % Replace by \answerYes{}, \answerNo{}, or \answerNA{}.
    \item[] Justification: Abstract and \S\ref{sec:intro} state three benchmark-level findings that match \S\ref{sec:experiments}: (i) cross-entity difficulty asymmetry between company and building tracks; (ii) the in-distribution to out-of-distribution gap dominates within-model variation, with TabPFN~v2 paired-bootstrap-significantly beating every tuned tree on 3/6 building-grouped tiers (Table~\ref{tab:app_t2_paired}); (iii) multimodal Sentinel-2+Clay embeddings help cross-city transfer but not in-distribution prediction (Figure~\ref{fig:t2_cross_city}, \S\ref{sec:exp_t2_s2}).

\item {\bf Limitations}
    \item[] Question: Does the paper discuss the limitations of the work performed by the authors?
    \item[] Answer: \answerYes{} % Replace by \answerYes{}, \answerNo{}, or \answerNA{}.
    \item[] Justification: Limitations are discussed in Appendix~\ref{app:limitations}, including data-coverage skew toward cities with disclosure laws, address-level geocoding for LA and Denver, the short five-year forecasting panel, and the late-concatenation S2 fusion design.

\item {\bf Theory assumptions and proofs}
    \item[] Question: For each theoretical result, does the paper provide the full set of assumptions and a complete (and correct) proof?
    \item[] Answer: \answerNA{} % Replace by \answerYes{}, \answerNo{}, or \answerNA{}.
    \item[] Justification: The paper introduces a benchmark and reports empirical results; it contains no theoretical results requiring proofs.

    \item {\bf Experimental result reproducibility}
    \item[] Question: Does the paper fully disclose all the information needed to reproduce the main experimental results of the paper to the extent that it affects the main claims and/or conclusions of the paper (regardless of whether the code and data are provided or not)?
    \item[] Answer: \answerYes{} % Replace by \answerYes{}, \answerNo{}, or \answerNA{}.
    \item[] Justification: All splits, hyperparameter search protocols, multi-seed configurations, and tuned best configs are released in Appendix~\ref{app:protocol} and Appendix~\ref{app:hp_configs}, alongside the code repository and Croissant-described dataset on Zenodo (DOI 10.5281/zenodo.20006582).

\item {\bf Open access to data and code}
    \item[] Question: Does the paper provide open access to the data and code, with sufficient instructions to faithfully reproduce the main experimental results, as described in supplemental material?
    \item[] Answer: \answerYes{} % Replace by \answerYes{}, \answerNo{}, or \answerNA{}.
    \item[] Justification: Code is released at \url{https://anonymous.4open.science/r/ana_review-7D14}; the building track is published as a Zenodo dataset (DOI 10.5281/zenodo.20006582) under CC-BY-4.0 with a Croissant 1.0 + RAI metadata file; the company track ships as a one-command reconstruction script (\S\ref{sec:dataset_t1}) against a free public API. A self-contained quickstart kit reproduces the headline TabPFN~v2 result with one command.

\item {\bf Experimental setting/details}
    \item[] Question: Does the paper specify all the training and test details (e.g., data splits, hyperparameters, how they were chosen, type of optimizer) necessary to understand the results?
    \item[] Answer: \answerYes{} % Replace by \answerYes{}, \answerNo{}, or \answerNA{}.
    \item[] Justification: Hyperparameter search ranges and selected configs are listed in Appendix~\ref{app:hp_configs}; multi-seed protocol and paired-bootstrap construction are in Appendix~\ref{app:protocol}; split definitions (stratification, ratios, year cuts) are in Appendix~\ref{app:splits}.

\item {\bf Experiment statistical significance}
    \item[] Question: Does the paper report error bars suitably and correctly defined or other appropriate information about the statistical significance of the experiments?
    \item[] Answer: \answerYes{} % Replace by \answerYes{}, \answerNo{}, or \answerNA{}.
    \item[] Justification: Headline numbers are 5-seed mean$\pm$std on both tracks; model-ordering claims use 1000-resample paired bootstrap on shared test rows reporting $\Delta R^2$, 95\% CI, and two-sided $p$-values. The full statistical protocol is in \S\ref{sec:benchmark_stats} and Appendix~\ref{app:protocol}; per-pair tables in Appendix~\ref{app:paired}.

\item {\bf Experiments compute resources}
    \item[] Question: For each experiment, does the paper provide sufficient information on the computer resources (type of compute workers, memory, time of execution) needed to reproduce the experiments?
    \item[] Answer: \answerYes{} % Replace by \answerYes{}, \answerNo{}, or \answerNA{}.
    \item[] Justification: Appendix~\ref{app:compute} reports compute (RTX A5000 24GB GPUs and multi-core CPU), per-experiment runtimes, and total project compute, including disclosure of preliminary runs that did not appear in the paper.
    
\item {\bf Code of ethics}
    \item[] Question: Does the research conducted in the paper conform, in every respect, with the NeurIPS Code of Ethics \url{https://neurips.cc/public/EthicsGuidelines}?
    \item[] Answer: \answerYes{} % Replace by \answerYes{}, \answerNo{}, or \answerNA{}.
    \item[] Justification: The work uses only publicly disclosed building energy/GHG data and free-tier company emissions disclosures, contains no personally identifiable information, and follows the NeurIPS Code of Ethics.

\item {\bf Broader impacts}
    \item[] Question: Does the paper discuss both potential positive societal impacts and negative societal impacts of the work performed?
    \item[] Answer: \answerYes{} % Replace by \answerYes{}, \answerNo{}, or \answerNA{}.
    \item[] Justification: Positive impact: enabling open, comparable evaluation of carbon-emission prediction supports climate-policy and operations research that depended on paywalled disclosures. Negative impact: model errors could be misused to greenwash poor performers; we discuss this risk and the limits of self-reported emissions in Appendix~\ref{app:limitations}.
    
\item {\bf Safeguards}
    \item[] Question: Does the paper describe safeguards that have been put in place for responsible release of data or models that have a high risk for misuse (e.g., pre-trained language models, image generators, or scraped datasets)?
    \item[] Answer: \answerNA{} % Replace by \answerYes{}, \answerNo{}, or \answerNA{}.
    \item[] Justification: The released assets are tabular emission-disclosure records and Sentinel-2 surface-reflectance embeddings; they pose no high misuse risk (no generative models, no facial recognition, no scraped private content).

\item {\bf Licenses for existing assets}
    \item[] Question: Are the creators or original owners of assets (e.g., code, data, models), used in the paper, properly credited and are the license and terms of use explicitly mentioned and properly respected?
    \item[] Answer: \answerYes{} % Replace by \answerYes{}, \answerNo{}, or \answerNA{}.
    \item[] Justification: All upstream data sources are cited at first use (\S\ref{sec:dataset}, Appendix~\ref{app:data_sources}); each source's license and terms are noted in Appendix~\ref{app:data_sources} and respected (open municipal portals are redistributed; CDU is reconstructed not redistributed).

\item {\bf New assets}
    \item[] Question: Are new assets introduced in the paper well documented and is the documentation provided alongside the assets?
    \item[] Answer: \answerYes{} % Replace by \answerYes{}, \answerNo{}, or \answerNA{}.
    \item[] Justification: The released building-track dataset ships with a Croissant 1.0 + MLCommons RAI metadata file (14 RAI fields including dataCollection, dataBiases, dataLimitations, dataSocialImpact, hasSyntheticData, and dataReleaseMaintenancePlan), a datasheet (Appendix~\ref{app:datasheet}), per-file MD5 checksums, and a quickstart kit with expected-output transcript.

\item {\bf Crowdsourcing and research with human subjects}
    \item[] Question: For crowdsourcing experiments and research with human subjects, does the paper include the full text of instructions given to participants and screenshots, if applicable, as well as details about compensation (if any)? 
    \item[] Answer: \answerNA{} % Replace by \answerYes{}, \answerNo{}, or \answerNA{}.
    \item[] Justification: The paper does not involve crowdsourcing or research with human subjects.

\item {\bf Institutional review board (IRB) approvals or equivalent for research with human subjects}
    \item[] Question: Does the paper describe potential risks incurred by study participants, whether such risks were disclosed to the subjects, and whether Institutional Review Board (IRB) approvals (or an equivalent approval/review based on the requirements of your country or institution) were obtained?
    \item[] Answer: \answerNA{} % Replace by \answerYes{}, \answerNo{}, or \answerNA{}.
    \item[] Justification: The paper does not involve crowdsourcing or research with human subjects, so IRB approval is not applicable.

\item {\bf Declaration of LLM usage}
    \item[] Question: Does the paper describe the usage of LLMs if it is an important, original, or non-standard component of the core methods in this research? Note that if the LLM is used only for writing, editing, or formatting purposes and does \emph{not} impact the core methodology, scientific rigor, or originality of the research, declaration is not required.
    %this research? 
    \item[] Answer: \answerYes{} % Replace by \answerYes{}, \answerNo{}, or \answerNA{}.
    \item[] Justification: LLMs were used only for writing assistance on the manuscript (typo correction and prose polishing). They did not produce experimental results, design analyses, or generate text for the technical contributions.

\end{enumerate}

\end{document}